\documentclass{article} % For LaTeX2e
\usepackage{collas2024_conference,times}
\usepackage{easyReview}
\usepackage{graphicx}
\usepackage{subcaption}
\usepackage{algorithm}
\usepackage{algorithmic}

\newcommand{\ours}{PLP}
\newcommand{\ourst}{PLP-T}
\newcommand{\gama}{GAMA}
\newcommand{\cdil}{CDIL}
\newcommand{\cond}{Contextual}

\usepackage{amsmath,amsfonts,bm}

\def\eqref#1{equation~\ref{#1}}
\def\1{\bm{1}}

\DeclareMathAlphabet{\mathsfit}{\encodingdefault}{\sfdefault}{m}{sl}
\SetMathAlphabet{\mathsfit}{bold}{\encodingdefault}{\sfdefault}{bx}{n}

\usepackage{hyperref}
\hypersetup{
    colorlinks=true,
    linkcolor=red,
    filecolor=magenta,
    urlcolor=blue,
    citecolor=purple,
    pdftitle={Overleaf Example},
    pdfpagemode=FullScreen,
    }

\title{Cross-Domain Policy Transfer by Representation Alignment via Multi-Domain Behavioral Cloning}

\author{Hayato Watahiki\thanks{Equal contribution.} , 
Ryo Iwase\footnotemark[1]\textsuperscript{\enspace}\thanks{
Ryo Iwase and Ryosuke Unno have left the University of Tokyo.} , 
Ryosuke Unno\footnotemark[2] , 
Yoshimasa Tsuruoka \\
The University of Tokyo \\
\texttt{watahiki@logos.t.u-tokyo.ac.jp} \\
}

\collasfinalcopy % Uncomment for camera-ready version, but NOT for submission.

\begin{document}

\maketitle

\begin{abstract}
Transferring learned skills across diverse situations remains a fundamental challenge for autonomous agents, particularly when agents are not allowed to interact with an exact target setup.
While prior approaches have predominantly focused on learning domain translation, they often struggle with handling significant domain gaps or out-of-distribution tasks. 
In this paper, we present a simple approach for cross-domain policy transfer that learns a shared latent representation across domains and a common abstract policy on top of it. 
Our approach leverages multi-domain behavioral cloning on unaligned trajectories of proxy tasks and employs maximum mean discrepancy (MMD) as a regularization term to encourage cross-domain alignment. 
The MMD regularization better preserves structures of latent state distributions than commonly used domain-discriminative distribution matching, leading to higher transfer performance.
Moreover, our approach involves training only one multi-domain policy, which makes extension easier than existing methods.
Empirical evaluations demonstrate the efficacy of our method across various domain shifts, especially in scenarios where exact domain translation is challenging, such as cross-morphology or cross-viewpoint settings.
Our ablation studies further reveal that multi-domain behavioral cloning implicitly contributes to representation alignment alongside domain-adversarial regularization.
\end{abstract}

\section{Introduction}
Humans have an astonishing ability to learn skills in a highly transferable way.
Once we learn a route from home to the station, for example, 
we can get to the destination 
using various modes of transportation (e.g., walking, cycling, or driving) 
in different environments (e.g., on a map or in the real world),
disregarding irrelevant perturbations (e.g., weather, time, or traffic conditions).
We identify the underlying structural similarities across situations, 
perceive the world, and accumulate knowledge in our way of abstraction.
Such abstract knowledge can be readily employed in diverse similar situations.
However, it is not easy for autonomous agents.
Agents trained with reinforcement learning (RL) or imitation learning (IL) often struggle to 
transfer knowledge acquired in a specific situation to another.
This is because the learned policies are strongly tied to the representations obtained 
under a particular training configuration, 
which is not robust to changes in an agent or an environment.

Previous studies have attempted to address this problem through various approaches.
Domain randomization \citep{tobin2017domain,peng2018sim,andrychowicz2020learning} aims to learn a policy that is robust to environmental changes by utilizing multiple training domains. 
However, it is unable to handle significant domain gaps that go beyond the assumed domain distribution during training, such as drastically different observations or agent morphologies.
Numerous methods have been proposed to overcome such domain discrepancies.
Earlier approaches learn domain-invariant state representations for imitation using a temporally-aligned dataset across domains \citep{gupta2017learning,liu2018imitation}.
In cases when we cannot assume such temporal alignment, other approaches utilize an adversarial objective based on domain confusion \citep{stadie2017third,yin2022cross,franzmeyer2022learn} or cross-domain cycle-consistency \citep{zakka2022xirl}.
These methods require online interaction for adaptation in the target domain to refine a policy, limiting their applicability. 

Recently, a few methods have been proposed that do not necessitate online interaction for adaptation \citep{kim2020domain,zhang2021learning,raychaudhuri2021cross}. 
These methods find a cross-domain transformation through the adversarial cross-domain translation of states, actions, or transitions.
Although these approaches show promising results, we have found two challenges they face. 
First, the direct domain translation can be difficult to discover when the discrepancy between domains is not small. 
For instance, if one agent has no legs while the other agent has multiple legs,
we cannot expect a perfect cross-robot translation of information on how the agent walks.
Second, these methods rely on signals from adversarial generation and other factors on top of generated elements,
lacking a more stable and reliable source of cross-domain alignment.

\begin{figure}[tb]
    \centering
    \includegraphics[width=.85\textwidth]{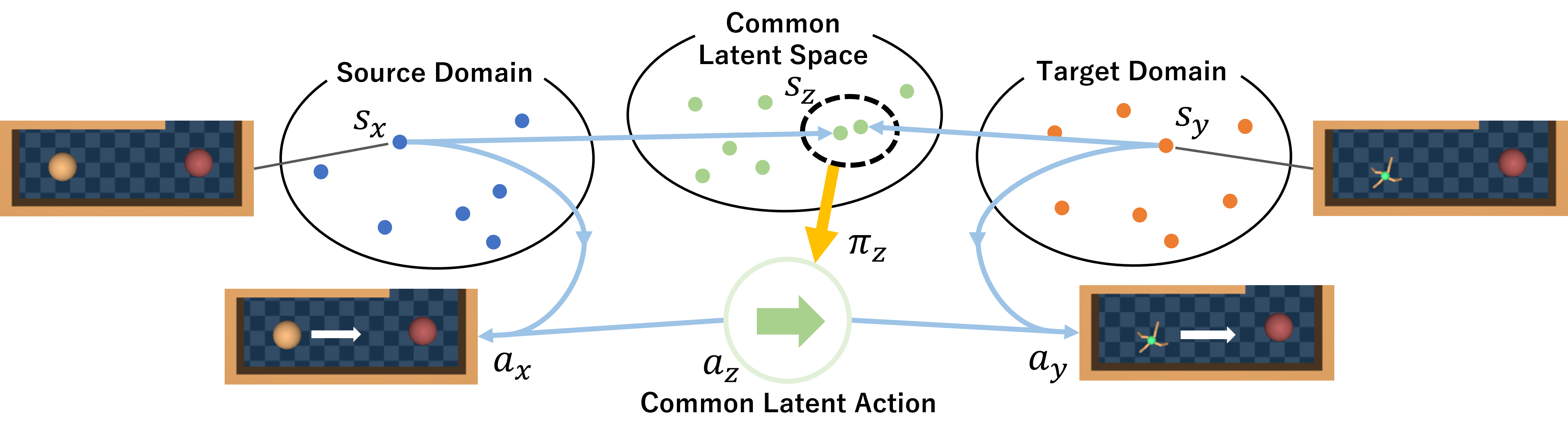}
    \caption{Illustration of a shared representation space and a common policy.
    We can transfer knowledge across domains if semantically similar states are mapped to close points in the shared latent space.
    The state projection onto the common latent space, the prediction of latent actions from the common state space (yellow arrow), and the decoding of latent actions to each domain 
    are modeled as a state encoder $\phi$, a common policy $\pi_z$, and an action decoder $\psi$ in our proposed method, respectively (c.f. Figure \ref{fig:overview}).
    }
    \label{fig:intro}
\end{figure}

In this work, we propose a method that does not rely on exact cross-domain correspondence and translation.
Our approach learns a shared latent representation across different domains and a common abstract policy on top of it (Figure \ref{fig:intro}).
After achieving the alignment, we can update a common policy for the target task using any learning algorithm with the mappings between the shared space and the original domains frozen. 
Combined with the frozen mappings, we can readily deploy the learned common policy in either domain without further online interaction.
As long as states fall within a distribution covered during the alignment process, 
the abstract policy can be continually updated in one domain to accept tasks performed on the learned feature space
in a transferrable manner without additional effort.
Similar to previous studies \citep{kim2020domain,zhang2021learning}, 
we assume access to a dataset of expert demonstrations of proxy tasks,
which are relatively simple tasks used for aligning representation.
In contrast to existing methods that stand on adversarial generation with domain confusion,
our approach leverages multi-domain behavioral cloning (BC) on proxy tasks as a core component for shaping a shared representation space.
We then add a few regularization terms on the latent state distributions to encourage cross-domain alignment.
Although adversarial generation with a domain classifier is a commonly used technique to match multiple distributions,
we observe that exact matching of distributions is overly demanding and sometimes disrupts the structure of a shared representation space. 
We instead employ maximum mean discrepancy (MMD) \citep{gretton2012kernel}, a widely utilized technique in domain adaptation \citep{long2013transfer,tzeng2014deep,baktashmotlagh2016distribution}.
We empirically confirm that it has a less detrimental impact on the representation structure.
We can optionally add more regularizations on the representation depending on proxy tasks. 
As an example, we add Temporal Cycle-Consistency learning (TCC)~\citep{dwibedi2019temporal} to promote state-to-state alignment using temporal information within a task rather than distribution overlap.
It is worth noting that our method only requires a policy network and a few loss terms, whereas other methods of offline cross-domain transfer usually require more models and objectives to optimize.
This enhances the robustness of our method against error accumulation, and more importantly, facilitates easy extensions of our approach depending on future requirements.

We evaluate our approach under various domain shifts, including changes in observation, action, viewpoint, and agent morphology. 
Our approach outperforms existing methods, particularly when discovering the exact domain translation of states or actions is challenging.
Moreover, our approach demonstrates superior adaptation capabilities to out-of-distribution tasks.
Additionally, we conduct extensive ablation studies to investigate the role of each loss term and show that BC is an indispensable component for representation alignment.
Perhaps surprisingly, our method exhibits some capability of cross-domain transfer even without regularization terms when the target task has similarity to proxy tasks. 
This implies that multi-domain BC can implicitly align latent representations across domains.

In summary, our main contributions are as follows:
\begin{itemize}
    \item We propose a method for cross-domain transfer that acquires a domain-shared feature space leveraging signals from multi-domain imitation in addition to domain confusion regularization with MMD, in contrast to the latest methods that rely on domain translation.
    \item We experimentally demonstrate the efficacy of our method under various domain shifts.
    Our method outperforms existing methods, especially in cross-robot transfer or cross-viewpoint transfer, where exact domain translation is hard to discover.
    \item We perform ablations to investigate the effect of each loss component for different domain shifts 
    and showcase the critical contribution of multi-domain BC to cross-domain representation alignment. 
    We also confirm that MMD regularization performs better than domain discriminative loss when a domain gap is not small.
\end{itemize}

\section{Related Work}
\paragraph{Unsupervised Domain Adaptation \& Distribution Matching}
Domain adaptation is a broad field aiming to transfer knowledge acquired from a source domain with ample labels to a target domain with sparse or no labeling. 
One approach is to match feature distributions of the two domains to maximize domain confusion.
Similar strategies have been explored in the context of generative adversarial networks (GANs), as these models aim to resemble data distributions.
The MMD has been a common approach since early work \citep{long2013transfer,tzeng2014deep,long2015learning,wang2018improving,wang2021rethinking},
while GAN-like domain discriminative loss serves as another popular measure for distribution matching \citep{ganin2016domain,tzeng2017adversarial}.
Other types of metrics have also been explored, such as correlations \citep{sun2016return,sun2016deep} or optimal transport \citep{courty2016optimal} between the distributions of domains.
In the context of generative models, \citet{li2015generative,dziugaite2015training} introduce MMD as an alternative to the standard Jensen-Shanon divergence.
\citet{arjovsky2017wasserstein,liu2017approximation} compare several divergence metrics for adversarial training and discuss properties of approximation, optimization, and convergence.

\paragraph{Unsupervised Domain Adaptation \& Correspondence Learning}
Another approach for domain adaptation is to learn domain translation.
CycleGAN \citep{Zhu_2017_ICCV} finds a cross-domain translation by generating the corresponding instances in another domain. 
If the output space is consistent between domains, we can enforce invariance on downstream components before and after the translation \citep{hoffman2018cycada,rao2020rl}.
Additionally, temporal relationships between frames \citep{sermanet2018time,dwibedi2019temporal}, cycle-consistency in agent trajectories \citep{zhang2021learning,wang2022weakly}, and optimal transport methods \citep{fickinger2022crossdomain} can be exploited to acquire domain translation or domain-invariant features. These features can subsequently be used for reward shaping in cross-domain imitation \citep{zakka2022xirl}.

\paragraph{Cross-Domain Policy Transfer between MDPs} 
Transferring a learned policy to a different environment is a long-standing challenge in policy learning. Most of the previous methods acquire some cross-domain metric to optimize and train a policy for a target task using a standard RL algorithm \citep{gupta2017learning,liu2018imitation,Liu2020State,zakka2022xirl,fickinger2022crossdomain} or a Generative Adversarial Imitation Learning (GAIL) \citep{ho2016generative}-based approach \citep{stadie2017third,yin2022cross,franzmeyer2022learn} through online interaction with the environment.
An adversarial objective based on domain confusion is typically used to match the state distribution of multiple domains.
In the reward calculation for standard RL, the distance between temporally-corresponding states \citep{gupta2017learning,liu2018imitation} or the distance from the goal \citep{zakka2022xirl} in the latent space is often used.
Similar to our method, some recent approaches do not assume online interaction for the adaptation to the target task.
\citet{kim2020domain,zhang2021learning,raychaudhuri2021cross} learn mappings between domains by adversarial generation of transitions or by CycleGAN,
while \citet{zhang2020invariant} impose domain confusion on its state representation to address domain shift in observation.
Our approach predicts actions without learning cross-domain mappings and focuses only on the shared structure, and also utilizes multi-domain BC for the representation alignment. 
In the experiment, we show that our approach can handle larger and more diverse domain shifts despite the simplicity compared to other baselines.
For cross-robot transfer, \citet{hejna2020hierarchically} train a portable high-level policy by using a subgoal position as a cross-robot feature.
\citet{gupta2022metamorph} cover a morphology distribution to generalize to unseen robots. We intend to perform direct policy transfer without making domain-specific assumptions.
Recently, large-scale BC that involves multiple domains and robots has shown impressive performance in domain generalization \citep{brohan2022rt,padalkar2023open,team2024octo}
and reported some degree of capability in cross-domain knowledge transfer.
The results shown in this paper are in line with such successes 
and provide an interpretation of what happens inside foundation policy models from the perspective of domain adaptation via representation alignment.

% Relationship between DeepMDP and the use of BC?
\paragraph{State Abstraction for Transfer}
Theoretical aspects of latent state representation have been analyzed in previous studies.
There exist several principled methods of state representation learning for transfer such as bisimulation \citep{castro2010using} and successor features \citep{barreto2017successor}.
\citet{gelada2019deepmdp} proved that the quality of a value function is guaranteed if the representation is sufficient to predict the reward and dynamics of the original Markov decision process (MDP).
In a similar context, \citet{zhang2020invariant,sun2022transfer} provide performance guarantees in multi-task settings or cross-domain transfer. 

\section{Problem Formulation}
\label{sec:formulation}

We consider a Markov decision process (MDP): $\mathcal{M} = (\mathcal{S}, \mathcal{A}, R, T)$, 
where $\mathcal{S}$ is a state space, $\mathcal{A}$ is an action space, $R: \mathcal{S} \times \mathcal{A} \rightarrow \mathbb{R}$ is a reward function, 
and $T: \mathcal{S}\times \mathcal{A} \times \mathcal{S} \rightarrow \mathbb{R}_{\geq 0}$ is a transition function.
We also define domain $d$ as a tuple $(\mathcal{S}_d$, $\mathcal{A}_d$, $T_d)$ and denote an MDP in domain $d$ as $\mathcal{M}_d: (\mathcal{S}_d, \mathcal{A}_d, R_d, T_d)$.
The aim of this paper is to transfer knowledge of a source MDP $\mathcal{M}_x$ in a source domain $x$ to a target MDP $\mathcal{M}_y$ in a target domain $y$. 
Here we assume that these MDPs share a common latent structure which is also an MDP $\mathcal{M}_z$.
Formally, we assume the existence of state mappings $\phi_x: \mathcal{S}_x\rightarrow\mathcal{S}_z, \phi_y: \mathcal{S}_y\rightarrow\mathcal{S}_z$ 
and action mappings $\xi_x:\mathcal{A}_x \rightarrow \mathcal{A}_z,$ $\xi_y:\mathcal{A}_y \rightarrow \mathcal{A}_z$
which translate states $s_x, s_y$ or actions $a_x, a_y$ in a domain 
into shared states $s_z$ or actions $a_z$, respectively, satisfying $T_z(\phi_d(s_d), \xi_d(a_d),\phi_d(s_d')) = T_d(s_d, a_d, s_d')$ and $R_z(\phi_d(s_d), \xi_d(a_d)) = R_d(s_d, a_d)$ for all $s_d, a_d, s_d'$ in each domain $d$.
In short, we assume that the common latent MDP is expressive enough to reproduce the dynamics and reward structure of both MDPs.
Regarding action mappings, instead of encoding actions, we mainly consider a reversed version of functions
$\psi_x: \mathcal{A}_z\times\mathcal{S}_x \rightarrow \mathcal{A}_x$ and 
$\psi_y: \mathcal{A}_z\times\mathcal{S}_y \rightarrow \mathcal{A}_y$ for end-to-end learning. These functions should receive raw domain-specific information $s_d$ to decode latent actions $a_d$.

Our goal is to learn the state mapping functions $\phi_x, \phi_y$ and the action mapping function $\xi_x, \xi_y$ so that any policy learned in the common latent space 
$\pi_z(a_z|s_z): \mathcal{S}_z\times\mathcal{A}_z\rightarrow\mathbb{R}_{\geq 0}$ can be immediately used in either MDP combined with the obtained mappings. 
In this paper, we use a deterministic policy and denote the latent policy as $\pi_z(s_z): \mathcal{S}_z\rightarrow\mathcal{A}_z$, although we can easily extend it to a stochastic policy.
We learn these mappings using expert demonstrations of \textit{proxy tasks} $\mathcal{K}$, which are simple tasks where we can easily collect demonstrations: 
$\mathcal{D} = \{(\mathcal{D}_{x,k}, \mathcal{D}_{y,k})\}_{k=1}^{|\mathcal{K}|}$, 
where $\mathcal{D}_{d, k} = \{\tau_{d, k, i}\}_{i=1}^{N}$ is a dataset of $N$ state-action trajectories $\tau_{d,k,i}$ of an expert in domain $d$, task $k$.
After we learn the relationships, we update the policy for a novel target task $k' \not\in\mathcal{K}$ in the source domain $x$, and finally evaluate its performance in the target domain $y$.

\section{Learning Common Policy via Representation Alignment}
\begin{figure}[tb]
    \begin{minipage}{.47\textwidth}
        \centering
        \includegraphics[width=.95\textwidth]{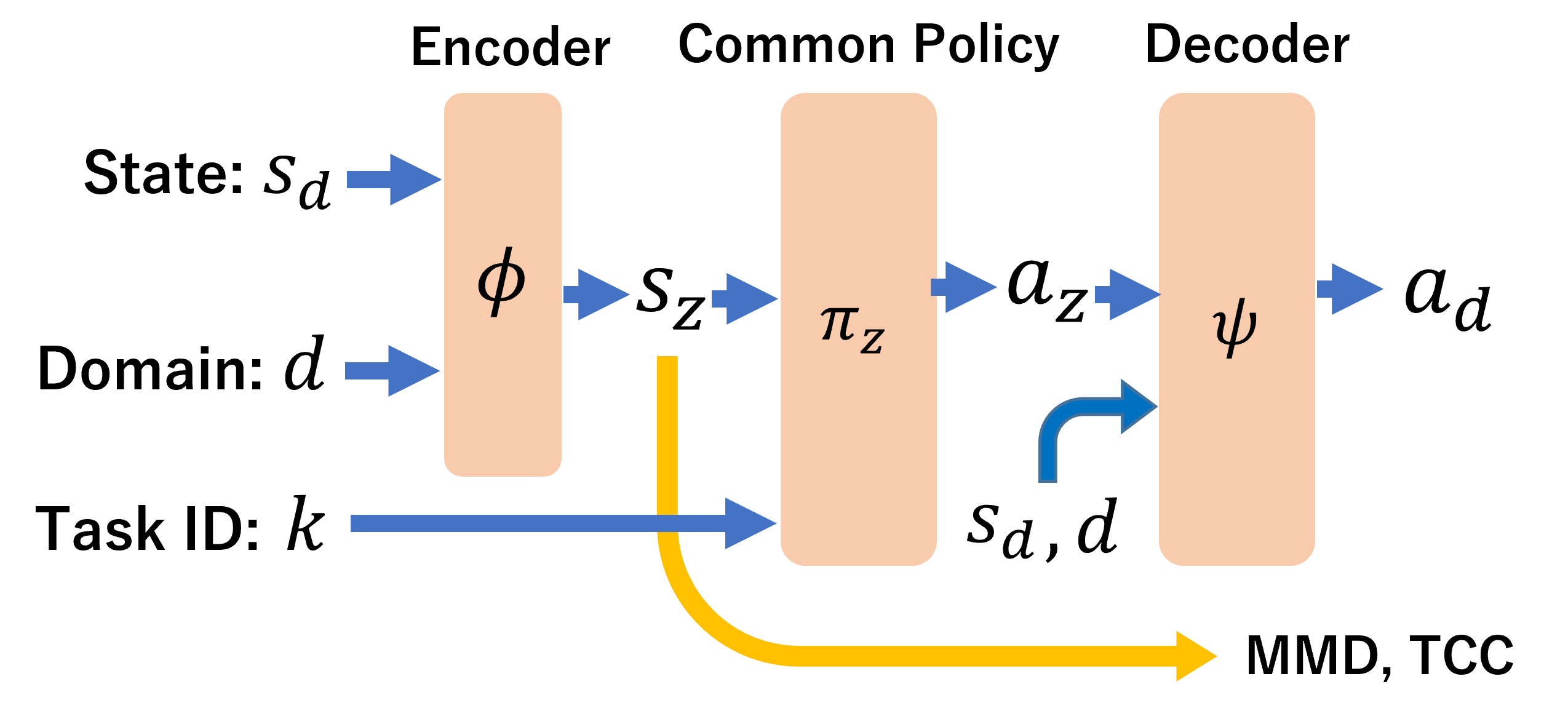} 
        \subcaption{Alignment phase. All modules are trainable.}
        \label{fig:align_phase}
    \end{minipage}
    \begin{minipage}{0.52\textwidth}
        \centering
        \includegraphics[width=.85\textwidth]{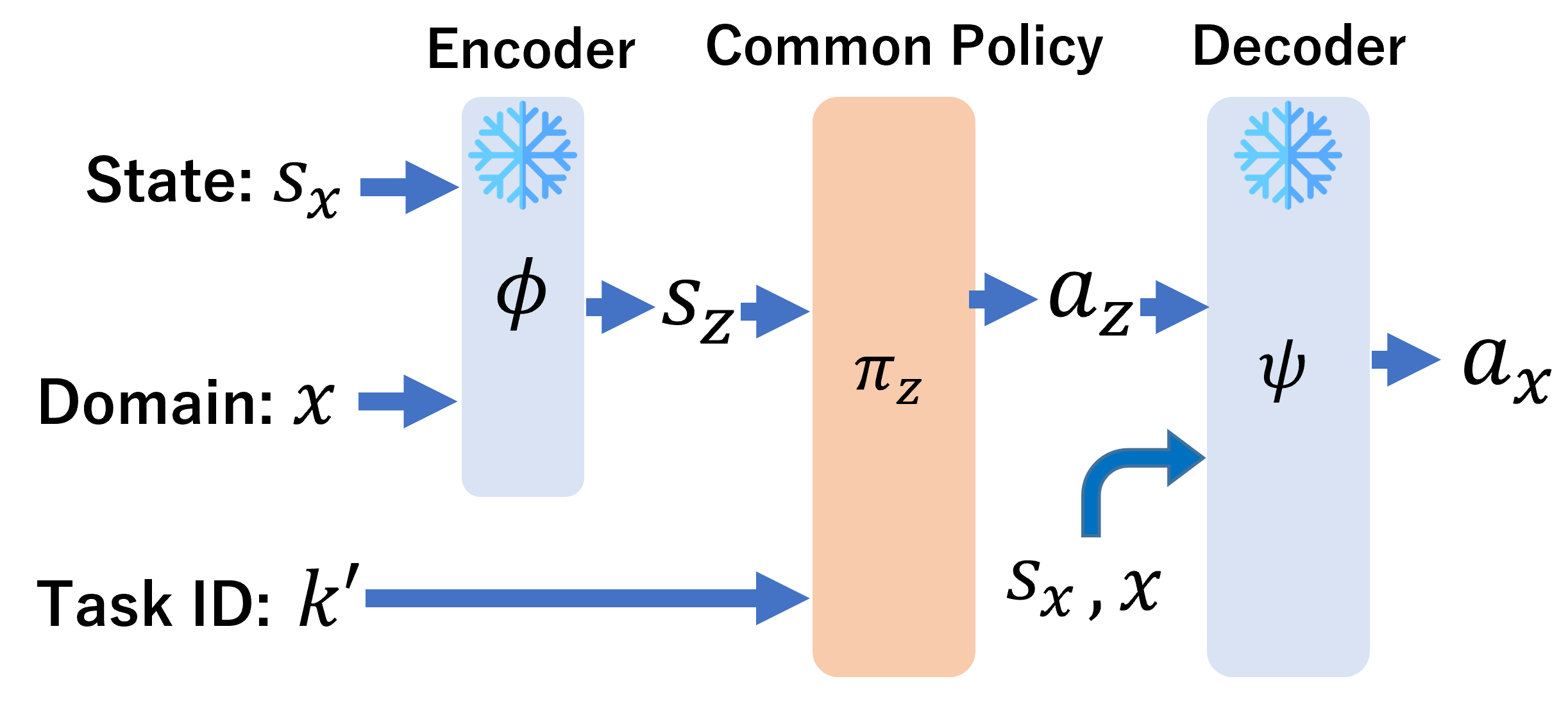} \subcaption{Adaptation phase. Only common policy is updated.} \label{fig:adapt_phase}
    \end{minipage}
    \caption{Overview of the training and inference procedure of our method. 
    (a) In the alignment phase, we train all modules of the policy via BC and regularization terms using trajectories of proxy tasks to obtain cross-domain representation alignment
    of latent states $s_z$.
    (b) In the adaptation phase, we only update the common policy built on a fixed representation space $s_z$ to adapt to the target task in the source domain. In inference, we can use the updated policy combined with the encoder and decoder already trained in the alignment phase.
    }
    \label{fig:overview}
\end{figure}

In this work, we aim to learn state mapping functions $\phi_x,\phi_y$, and action mapping functions $\xi_x, \xi_y$ or equivalents,
and use them to transfer the policy learned in one domain to another.
Our algorithm consists of two steps as illustrated in Figure \ref{fig:overview}: 
(i) Cross-domain representation alignment, (ii) Policy adaptation to a target task in the source domain. 
We call them the {\it alignment} phase and the {\it adaptation} phase, respectively.
After the adaptation phase, the learned policy of a target task can be readily used in the target domain 
without any fine-tuning or further interaction with the target domain \citep{gupta2017learning,liu2018imitation,zakka2022xirl,fickinger2022crossdomain,yin2022cross,franzmeyer2022learn},
or a policy learning in the mapped target domain \citep{raychaudhuri2021cross}.

\subsection{Cross-Domain Representation Alignment}
\label{sec:explanation_alignment}
In the alignment phase, we aim to learn the state and action mappings and acquire a domain-shared feature space that can be used in either the source domain or the target domain.
We represent our policy as a simple feed-forward neural network as shown in Figure \ref{fig:overview}. It consists of three components: a state encoder, a common policy, and an action decoder.
They correspond to $\phi(s)$, $\pi(s_z)$, and $\psi(a_z)$, respectively.
We feed domain ID $d$ to the encoder and the decoder, and one-hot encoded task ID $k$ to the common policy,
instead of using separate networks for each domain, to handle multiple domains and tasks with a single model.

\begin{figure*}[tb]
    \centering
    \begin{minipage}{0.33\textwidth}
        \centering
        \includegraphics[width=.75\textwidth]{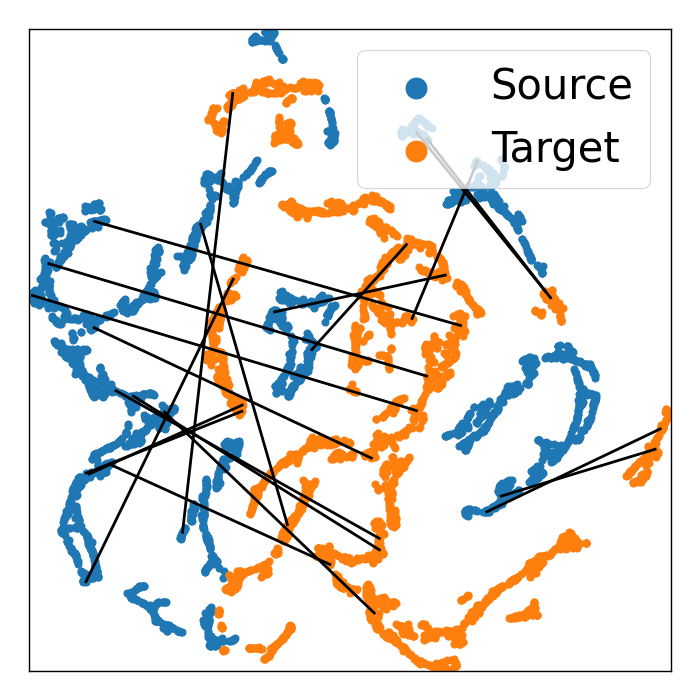}
        \subcaption{P2P, BC Only}
        \label{fig:dist_p2p_bconly}
    \end{minipage}
    \begin{minipage}{0.33\textwidth}
        \centering
        \includegraphics[width=.75\textwidth]{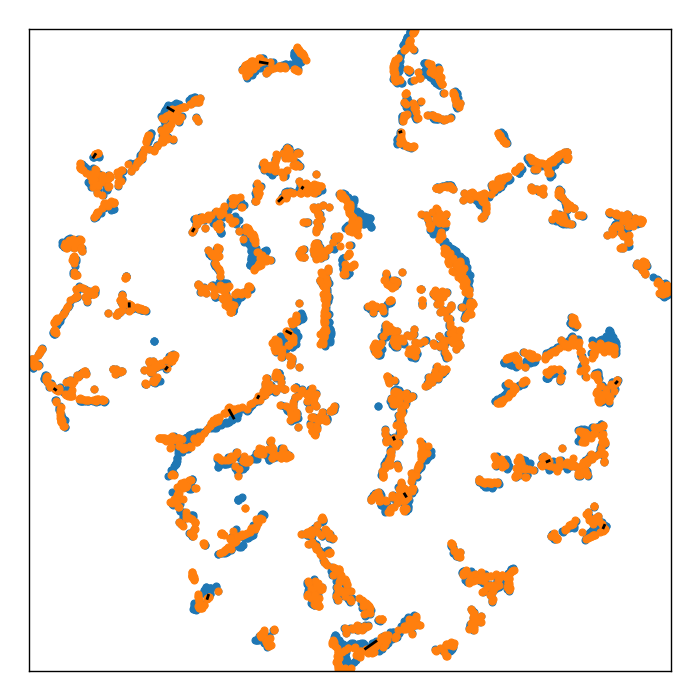}
        \subcaption{P2P, BC + MMD}
        \label{fig:dist_p2p_mmd}
    \end{minipage}
    \begin{minipage}{0.32\textwidth}
        \centering
        \includegraphics[width=.75\textwidth]{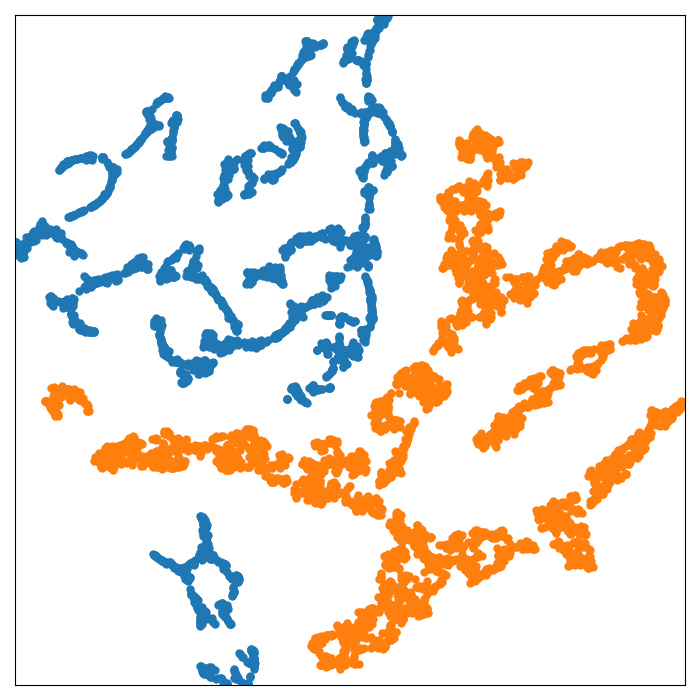}
        \subcaption{P2A, BC Only }
        \label{fig:dist_p2a_bconly}
    \end{minipage}
    \begin{minipage}{0.33\textwidth}
        \centering
        \includegraphics[width=.75\textwidth]{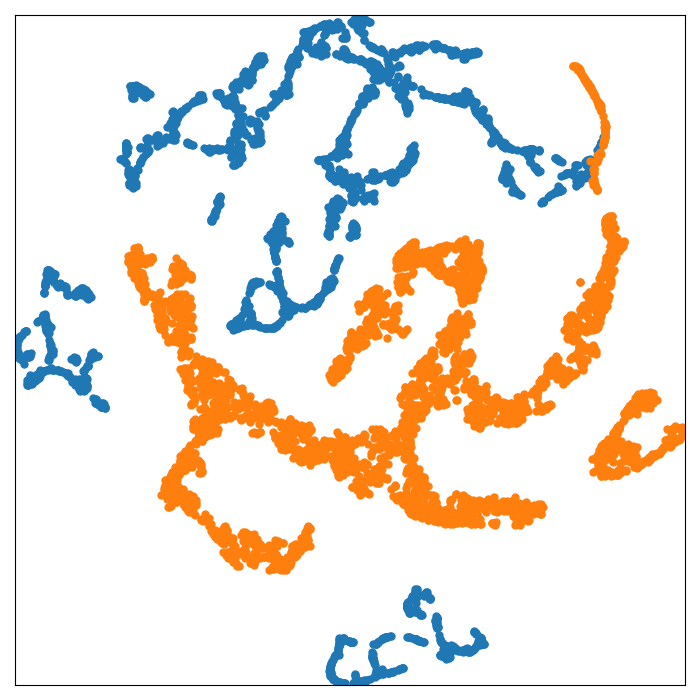}
        \subcaption{P2A, BC + MMD }
        \label{fig:dist_p2a_mmd}
    \end{minipage}
    \begin{minipage}{0.33\textwidth}
        \centering
        \includegraphics[width=.75\textwidth]{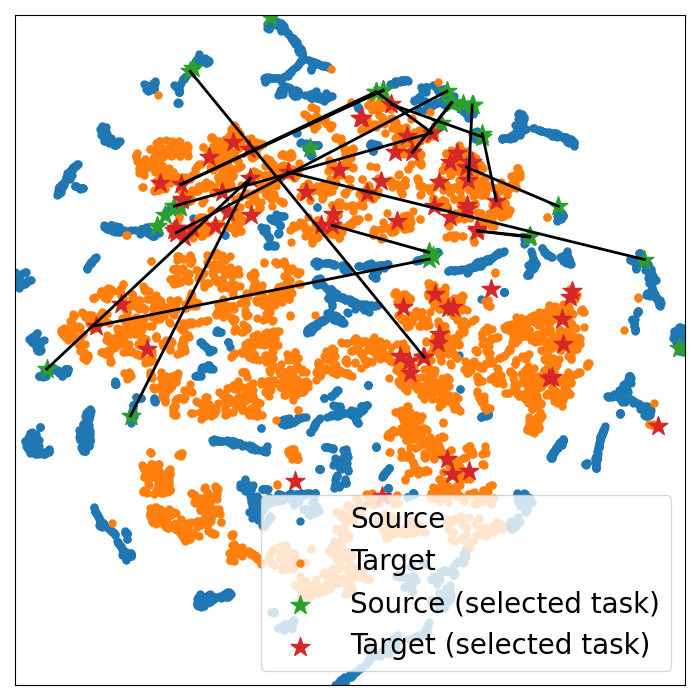}
        \subcaption{P2A, BC + MMD + TCC}
        \label{fig:dist_p2a_tcc}
    \end{minipage}
    \begin{minipage}{0.32\textwidth}
        \centering
        \includegraphics[width=.75\textwidth]{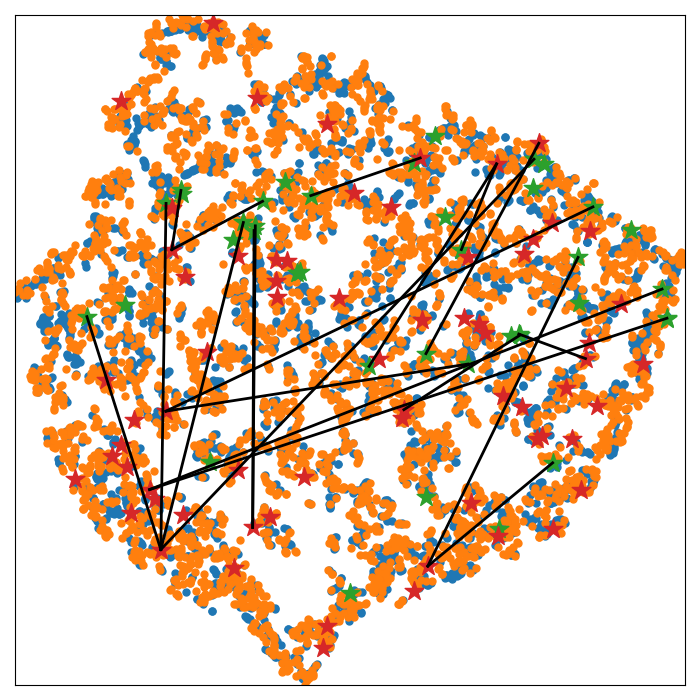}
        \subcaption{P2A, with discriminative}
        \label{fig:dist_p2a_adv}
    \end{minipage}
    \caption{Effect of each objective on the latent state distribution $(s_z)$. The representations are projected to 2D space by t-SNE \citep{van2008visualizing}. 
    Here we sample corresponding states from two domains of the Maze environment in our experiment (c.f. Section \ref{sec:env}). 
    (a-b) A plot for P2P-medium environment, where the two domains are the same except for the format of states and actions. 
    Black lines connect 20 corresponding state pairs. BC shapes the latent space in each domain separately (\ref{fig:dist_p2p_bconly}), and
    the MMD loss encourages the distribution alignment. (\ref{fig:dist_p2p_mmd}). 
    (c-f) A plot for P2A-medium environment, where two domains have a large discrepancy. 
    In \ref{fig:dist_p2a_tcc} and \ref{fig:dist_p2a_adv}, we additionally plot states in a specific task and connect 20 corresponding state pairs based on 2D positions in the maze.
    With the help of MMD and TCC, our method obtains a shared latent space keeping the original structure (\ref{fig:dist_p2a_tcc}). If we use discriminative loss instead of MMD, the latent structure can be disrupted and states are scattered across the entire space (\ref{fig:dist_p2a_adv}).
    }
    \label{fig:repr}
\end{figure*}

We train the network via multi-domain BC with the MMD \citep{gretton2012kernel} loss and optionally with TCC~\citep{dwibedi2019temporal}. 
Each objective plays a different role in shaping the aligned representation space.
Through BC, the model learns a relationship between raw states and corresponding expert actions. We simply calculate the L2 loss on raw actions as follows:
\begin{equation}
    \displaystyle \mathcal{L}_\text{BC} = \mathbb{E}_{(s_d, a_d, d, k) \sim \mathcal{D}}
    \left[\|\psi_d(\pi_z(\phi_d(s_d), k), s_d) - a_d \|^2 \right].  \label{eq:bc}
\end{equation}
If we na\"ively optimize this objective with trajectories from both domains, the model can learn a separate state representation in each domain.
If temporally-aligned trajectories are available
(i.e. $(\phi_x(s_x^t), \psi_x(a_x^t)) = (\xi_y(s_y^t), \xi_y(a_y^t))$ at timestep $t$),
we can directly make corresponding representations $\phi_x(s_x),\phi_y(s_y)$ or $\xi_x(a_x),\xi_y(a_y)$ close together \citep{gupta2017learning}. 
However, we do not assume such an alignment and thus need to use other regularizations to align the representations.

MMD is a non-parametric metric that compares the discrepancy between two distributions based on two sets of data points. 
Distribution matching via MMD minimization is widely used for GANs~\citep{li2015generative,dziugaite2015training}
and domain adaptation~\citep{long2013transfer,tzeng2014deep,long2015learning,wang2018improving,wang2021rethinking}
in computer vision,
and is also used to align the support of two action distributions~\citep{kumar2019stabilizing}.
We apply it to latent states encoded in the source domain and target domain:
\begin{align}
\mathcal{L}_\text{MMD} = \mathbb{E}_{s_x, s'_x\sim \mathcal{D}_x}[f(\phi_x(s_x), \phi_x(s'_x))] 
&+\mathbb{E}_{s_y, s'_y\sim \mathcal{D}_y}[f(\phi_y(s_y), \phi_y(s'_y))]  \notag \\ 
&- 2\mathbb{E}_{s_x\sim \mathcal{D}_x, s_y\sim \mathcal{D}_y}[f(\phi_x(s_x), \phi_y(s_y))], \label{eq:mmd}
\end{align}
where $\mathcal{D}_d$ is a dataset of domain $d$ and $f$ is a kernel function that measures the similarity between sets of data points.
We adopt the Gaussian kernel as $f$, combined with batch-level distance normalization to avoid representation corruption. 
With Gaussian kernel, minimizing the MMD corresponds to matching all moments of two distributions \citep{li2015generative}.
See Appendix \ref{app:plp_detail} for the details.
For a similar purpose, domain discriminative training, where we train state encoder or domain translation functions to fool the domain discriminator, 
is frequently used in the context of cross-domain policy transfer \citep{kim2020domain,zhang2020invariant,raychaudhuri2021cross,franzmeyer2022learn}. 
However, this objective enforces the complete match of the distributions even if the structure of latent space or state frequency differs.
Here our goal is not to achieve a perfect match of samples from different domains or complete removal of domain-specific information, 
but rather to achieve sufficient alignment to share the same path in a policy.
In this paper, we posit that the MMD is more suitable for this modest alignment.
From a theoretical perspective, \citet{liu2017approximation} proved that the Jensen-Shanon divergence, a standard objective of GANs, is \textit{stronger} than the MMD.
The paper further showed that the MMD belongs to the weakest family of divergence metrics for distribution approximation.
The weakness is actually a desirable property because a distance is more likely to be continuous over parameters of distributions, 
better handling disjoint supports, and thus providing more information for optimization \citep{arjovsky2017wasserstein}.
Figure \ref{fig:dist_p2a_tcc} and \ref{fig:dist_p2a_adv} show how the MMD and discriminative objective affect the alignment when a domain discrepancy is large. 
MMD encourages the distribution overlap modestly,
whereas discriminative training forces the exact match disregarding the original structure.
We also observed that, in visual input settings, aligning representations containing image embedding with the MMD is more stable than with discriminative training.
We evaluate the effect of this difference in our experiment.
% TODO: WGAN / Hausdorff.
\begin{algorithm}[t]
\caption{Representation Alignment}
\label{alg:alignment}
\begin{algorithmic}
  \REQUIRE State encoder $\phi(s_d, d)$, Common policy $\pi_z(s_z, k)$, Action decoder $\psi(a_z, s_d, d)$, Loss weights $\lambda_\text{MMD}, \lambda_\text{TCC}$.
  \REQUIRE Expert trajectories of proxy tasks $\mathcal{D} = \{(s_d, a_d, d, k)\}$.
  \STATE
  \WHILE{not converged} 
  \STATE Sample a batch of state action pairs: $\{(s_d, a_d, d, k)\} \sim \mathcal{D}$.
  \STATE Calculate $\mathcal{L}_\text{BC}$ in (\ref{eq:bc}) and $\mathcal{L}_\text{MMD}$ in (\ref{eq:mmd}) on the batch.
  \STATE [Optional] Calculate $\mathcal{L}_\text{TCC}$ on the batch if $\lambda_\text{TCC} > 0$.
  \STATE Calculate total loss according to (\ref{eq:final}):
    $\mathcal{L}_\text{align} = \mathcal{L}_\text{BC} + \lambda_\text{MMD} \mathcal{L}_\text{MMD} + \lambda_\text{TCC} \mathcal{L}_\text{TCC}$.
  \STATE Update model parameters of $\phi, \pi_z, \psi$ to minimize $\mathcal{L}_\text{align}$.
  \ENDWHILE
\end{algorithmic}
\end{algorithm}

We optionally impose temporal alignment regularization on the representation $s_z$ using TCC.
Given two trajectories of the same task, TCC enforces cycle consistency between corresponding frames in the latent space. 
TCC only selects the frame, in contrast to CycleGAN~\citep{Zhu_2017_ICCV} generating the counterpart, which makes the learning easier.
Specifically, we randomly sample two state trajectories of the same task from both domains and encode each frame with the state encoder $\phi_d(s_d)$. 
Here we denote the two encoded trajectories as $U = (u_1, \cdots u_{|U|})$, $V = (v_1, \cdots v_{|V|})$, 
where $u_t = \phi_{d_1}(s_{d_1}^t)$, $v_t = \phi_{d_2}(s_{d_2}^t)$. 
For each state $u_i$ in $U$, we calculate the soft nearest-neighbor in $V$: $\tilde{v}^i = \sum_j^{|V|} \text{softmax}_j(-\|u_i - v_j\|)\cdot v_j$.
Then we choose a state in the first trajectory that is closest to $\tilde{v}^i$, and optimize the cross-entropy loss so that this sequence of mappings comes back to the original frame $u_i$.
The final TCC objective is $\mathcal{L}_\text{TCC} = - \sum_i^{|U|} \sum_k^{|U|} \mathbf{1}_{k=i} \log(y_k^i)$, where $y_k^i = \text{softmax}_k(-\|\tilde{v}^i - u_k\|^2) \label{eq:tcc}$, and $\mathbf{1}$ is an indicator function. 

Combining these three objectives, (\ref{eq:bc}), (\ref{eq:mmd}), and $\mathcal{L}_\text{TCC}$, we have our objective for the alignment phase:
\begin{align}
\min_{\phi, \pi_z, \psi} \mathcal{L}_\text{align} 
= \min_{\phi, \pi_z, \psi} \mathcal{L}_\text{BC} + \lambda_\text{MMD} \mathcal{L}_\text{MMD} + \lambda_\text{TCC} \mathcal{L}_\text{TCC}, \label{eq:final} 
\end{align}
where $\lambda_\text{MMD}$ and $\lambda_\text{TCC}$ are hyperparameters that define the importance of the regularization terms. 
The algorithm is summarized in Algorithm \ref{alg:alignment}.
Figure \ref{fig:repr} shows the effect of each loss term in our experiment. We discuss it more in Section \ref{sec:qualitative}.

It is worth mentioning that we should carefully design a proxy task dataset to meet several requirements for effective representation alignment.
i) Trajectories labeled as the same task must be mappable to similar trajectories in the latent space to satisfy the assumption described in Section \ref{sec:formulation}. 
Tasks should be solved similarly in both domains so that latent state distributions have a similar structure.
ii) Proxy tasks should cover a broad range of states enough to generalize to states in target tasks to adapt to, as we fix state representation after the alignment phase.
iii) If TCC is utilized, an agent should not visit the same state more than once as TCC requires unique correspondences for each state.

In some experimental setups, we observe that the state input for the action decoder can degrade the performance because the decoder can obtain all necessary information except task ID $k$ without the common policy. 
We can eliminate this effect by removing the state input for the decoder when the action prediction does not require domain-specific state information.

\subsection{Policy Adaptation}
\label{sec:adaptation}
In the adaptation phase, we update the common policy on top of the aligned latent space trained in the alignment phase. 
This adaptation can be solely in the source domain with any learning algorithm including reinforcement learning as long as the latent space is fixed. 
As described in Figure \ref{fig:adapt_phase}, we freeze weights of the encoder and decoder during the process.
In our experiments, we update the common policy by BC on expert trajectories in the source domain $\mathcal{D}_{x, k'}$:
\begin{align}
\min_{\pi_z}\mathcal{L}_\text{adapt} = \min_{\pi_z} \mathcal{L}_\text{BC}.  \label{eq:adapt}
\end{align}
Algorithm \ref{alg:adaptation} in Appendix \ref{app:algorithm} summarizes the procedure of BC-based adaptation.
When the discrepancy between domains is not small, or the alignment is imperfect, the update only with source domain data can make a common policy more or less domain-specific. 
This issue can be addressed by mixing the data used in the alignment phase with the data for the adaptation for the regularization.

\section{Experiments}

We conduct experiments to answer the following questions: 
(i) Can our method align latent states of different domains?
(ii) Can our method achieve zero-shot cross-domain transfer across various settings?
(iii) How does each loss contribute to transfer under different types of domain shifts?

\begin{figure}[tb]
    \centering
    \begin{minipage}{0.16\textwidth}
        \centering
        \includegraphics[width=.95\textwidth]{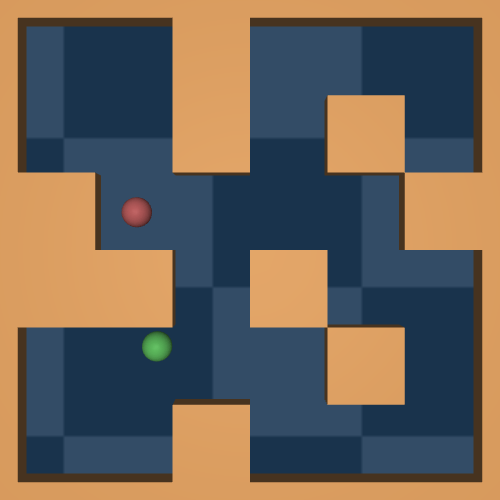}
        \subcaption{P2P-medium}
        \label{fig:maze2d}
    \end{minipage}
    \begin{minipage}{0.35\textwidth}
        \centering
        \includegraphics[width=.9\textwidth]{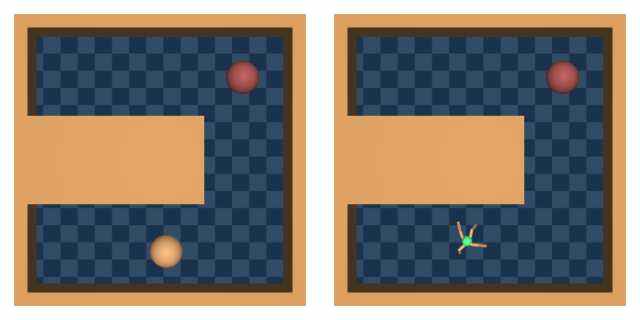}
        \subcaption{P2A-umaze}
        \label{fig:point-ant}
    \end{minipage}
    \begin{minipage}{0.35\textwidth}
        \centering
        \includegraphics[width=.9\textwidth]{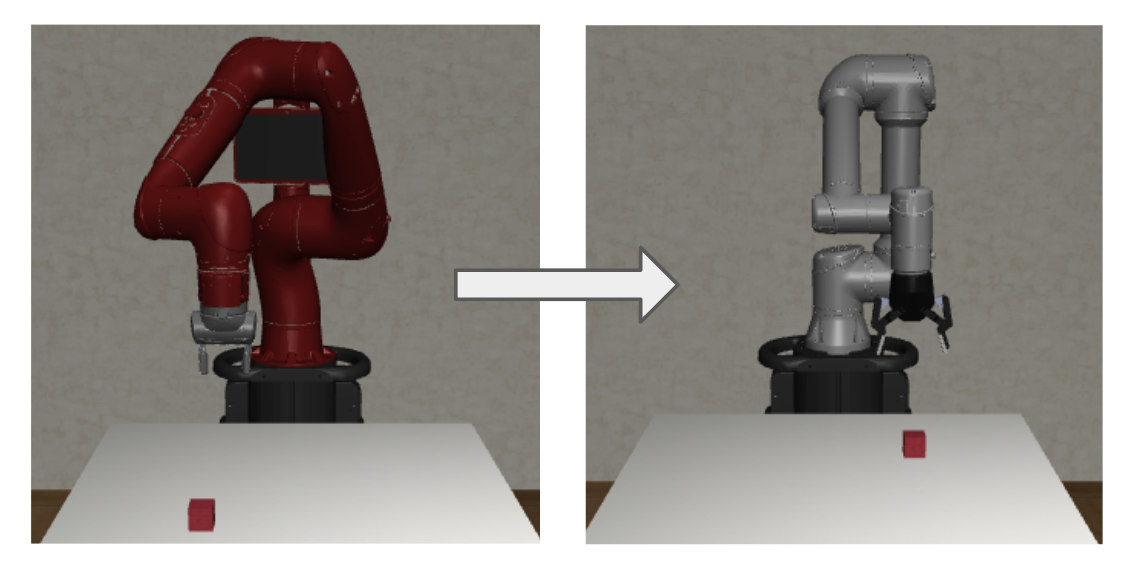}
        \subcaption{R2R-Lift}
        \label{fig:r2r}
    \end{minipage}
    \begin{minipage}{\textwidth}
        \centering
        \begin{minipage}{0.28\textwidth}
            \centering
            \includegraphics[width=.99\textwidth]{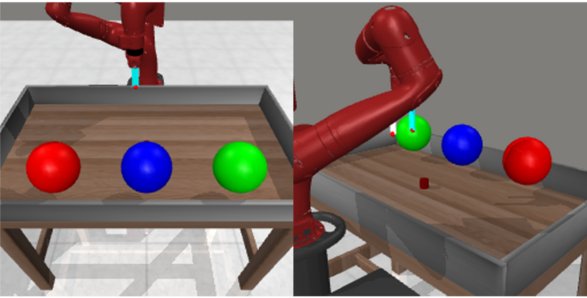}
            \subcaption{V2V-Reach}
            \label{fig:v2v_color}
        \end{minipage}
        \begin{minipage}{0.48\textwidth}
            \centering
            \includegraphics[width=.85\textwidth]{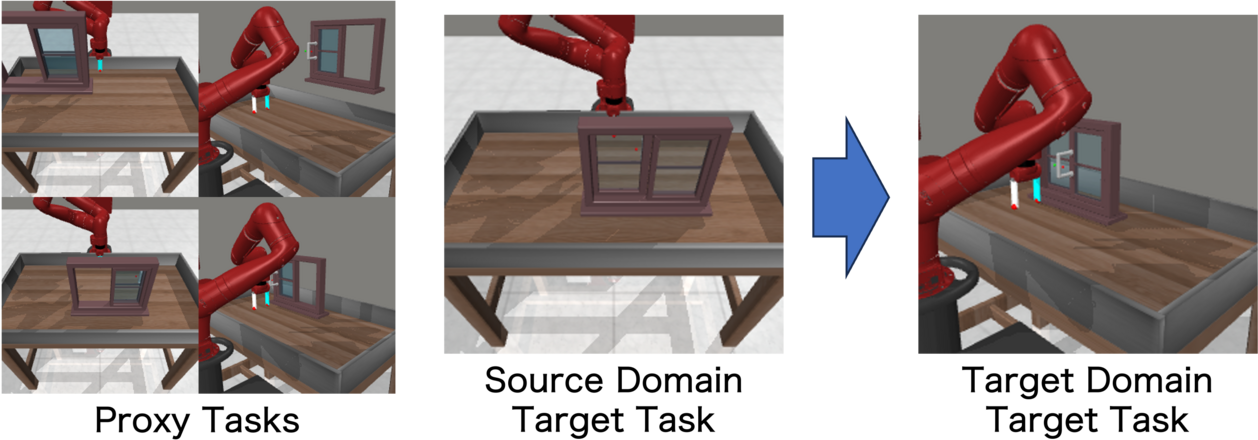}
            \subcaption{V2V-Open}
            \label{fig:v2v_open}
        \end{minipage}
        \begin{minipage}{0.22\textwidth}
            \centering
            \includegraphics[width=.65\textwidth]{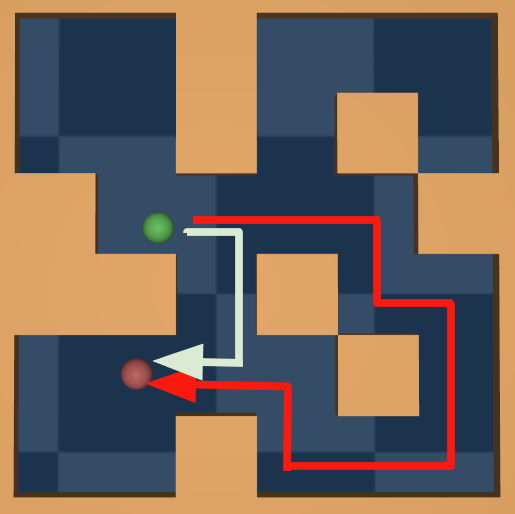}
            \subcaption{Target route of P2P-OOD}
            \label{fig:detour}
        \end{minipage}
    \end{minipage}
    \caption{(a-e) Pictures of P2P, P2A, R2R-Lift, and V2V. 
    The red points in the mazes (\ref{fig:maze2d}) and (\ref{fig:point-ant}) show the goals, which are not observable for the agents. (f) The route for OOD tasks. The red arrow shows the target route, 
    while the dataset only contains the shortest paths as the green arrow.}
    \label{fig:env}
\end{figure}

\subsection{Environments and Tasks}
\label{sec:env}
For locomotion tasks, we use the Maze environment of D4RL \citep{fu2020d4rl} (Figure \ref{fig:env}).
An agent explores two types of mazes, {\it umaze} and {\it medium}, toward a designated goal position.
An agent observes proprioceptive state vectors.
A task is defined as a combination of a starting area and a 2D position of a goal. 
Tasks with different goals from the one of a target task are used as proxy tasks.
In \textit{OOD} (Out-of-Distribution) variation, the target task is to take a detour as performed in given source domain demonstrations (Figure \ref{fig:detour}), despite the alignment dataset only containing shortest-path trajectories.
We create two setups with different domain shifts. 
(i) Point-to-Point (\textbf{P2P}): a Point agent learns from another Point agent with different observation and action spaces. 
The source agent receives x-y swapped observation and takes inverted action (i.e., multiplied by -1) of the target agent.
The two domains are essentially identical and thus we can calculate ground-truth correspondence between domains for evaluation.
For ablation, we create \textit{P2P-obs} for the medium maze, where we only keep the observation shift.
(ii) Point-to-Ant (\textbf{P2A}): an Ant agent learns from a Point agent, which has drastically different observations, actions, dynamics, and physical capabilities.

For manipulation, we create three tasks. 
(i) Robot-to-Robot Lift task (\textbf{R2R-Lift}) from robosuite \citep{robosuite2020}:
a robot has to learn from another robot with a different number of joints and different grippers 
to pick up a target block in a position unseen during the alignment phase (Figure \ref{fig:r2r}).
The observations are low-dimensional vectors with up to 37 dimensions. 
Both robots are controlled by delta values of a 3D position of the end effector and a 1D gripper state, 
although the outcome of the action can vary to some extent.
A single goal position in 3D space is selected as a target task and other goals that are not in the same height or same 2D position are used for proxy tasks.
In Viewpoint-to-Viewpoint (V2V) environments constructed based on environments in Meta-World \citep{yu2020meta}, 
a robot learns from demonstrations from a different viewpoint. 
The robot observes an RGB image from a specific viewpoint in addition to common proprioceptive inputs.
We use two setups. 
(ii) \textbf{V2V-Reach}: the robot needs to move its arm to a goal shown as a ball with a specific color in an image.
The order of balls is randomly initialized. We use a single color for a target task and use the rest for proxy tasks.
(iii) \textbf{V2V-Open}: the robot needs to open the window in a random position. The proxy tasks only contain trajectories of closing the window, where the robot moves its arm in the opposite direction.
For the details, please refer to Appendix \ref{app:env}.

\subsection{Baselines}
We refer to our method as \textbf{\ours}~(\textbf{P}ortable \textbf{L}atent \textbf{P}olicy). 
We denote the method with and without the TCC term as \textit{\ourst}~and \textit{\ours}, respectively.
We also create \textit{\ourst-disc} and \textit{\ours-disc} where we replace our MMD loss with the domain discriminative loss. 
For the reason mentioned in Section \ref{sec:explanation_alignment}, we do not provide states for the decoder in R2R and V2V. 
We compare our approach with the following baselines.
\textbf{\gama}~\citep{kim2020domain} learns direct cross-domain mappings of states and actions via adversarial training on generated transitions using a dynamics model. \gama~solves the target task using the updated source domain policy combined with the learned cross-domain translation. 
\textbf{\cdil}~\citep{raychaudhuri2021cross} learns a state translation function with CycleGAN~\citep{Zhu_2017_ICCV} for cross-domain transfer. \cdil~additionally employs information on task progression via regression of the progression,
which has a similar role to that of TCC. This method finally trains a policy for the target domain using translated states and actions inferred by an inverse dynamics model.
\textbf{\cond~policy} (\textit{\cond}~for short) is a policy with Transformer \citep{vaswani2017attention} architecture that takes source domain demonstration in the encoder and takes the observation history in the decoder, and outputs the next action.
It only has the alignment phase as it requires a pair of demonstrations from both domains for the training.
\textbf{BC} learns a flat policy that digests a state, domain ID, and task ID at once. It is trained with the same parameters as \ours.
See Appendix \ref{app:plp_detail} and \ref{app:baselines} for the details.

\begin{figure}[tb]
    \centering
    \includegraphics[width=.80\textwidth]{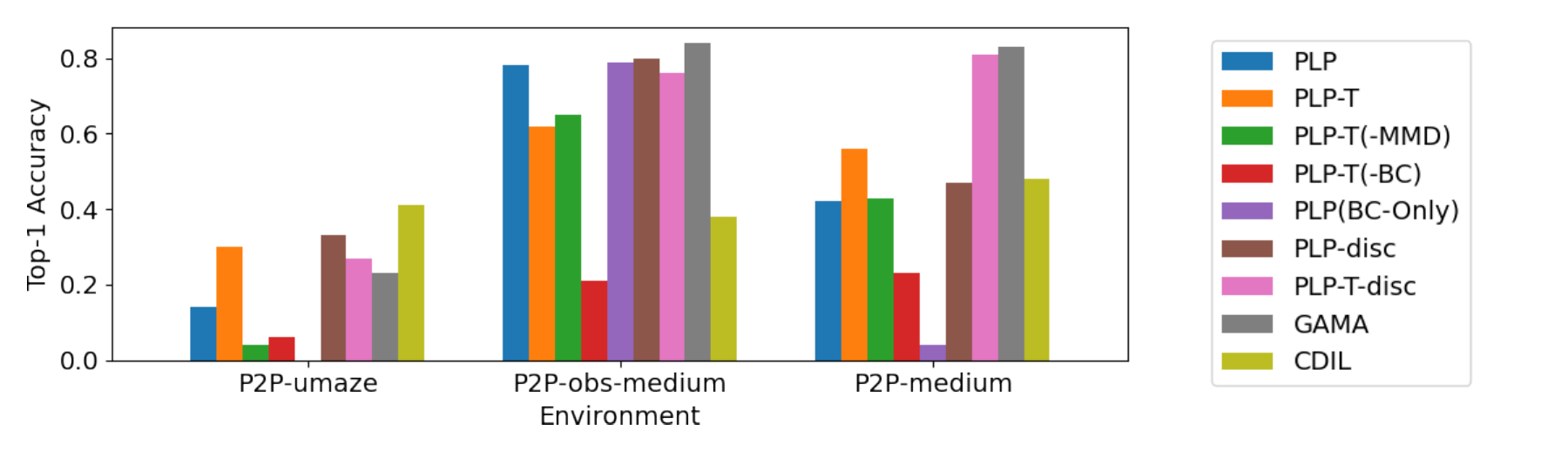}
    \caption{Alignment scores for P2P environments. The values are top-1 accuracies of finding a corresponding state from 1k random states based on the distance in the latent space (or state space for \gama).
    The regularization terms of \ours~enhance the alignment in P2P-umaze and P2P-medium, with BC contributing the most to the alignment.
    In P2P-obs, even the BC-Only version successfully aligns the representations.
    }
    \label{fig:align_score}
\end{figure}
\begin{figure}[tb]
    \centering
    \includegraphics[width=.90\textwidth]{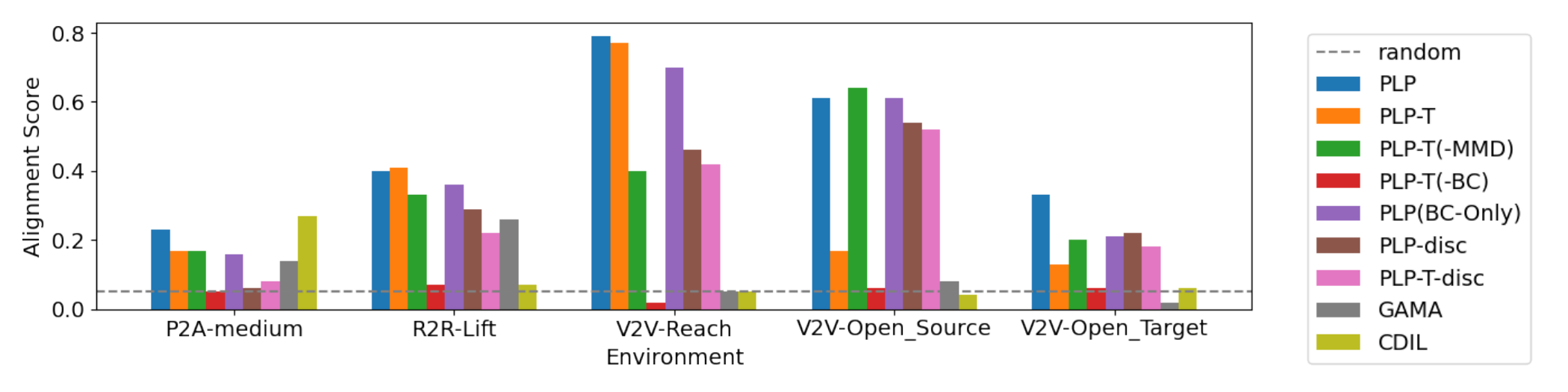}
    \caption{Alignment scores for P2A, R2R and V2V. The values are the probability of corresponding states having top-5\% closeness in the latent space (or state space for \gama) among all state pairs. 
    For V2V-Open, we calculate scores separately for states in the source task (Window-Close) used for the alignment and the target task (Window-Open).
    \ours~aligns the representations consistently well and the BC-Only variant achieves comparable performance in several environments.
    The performance of \ours-disc is lower than that of \ours~due to the restrictive regularization, 
    while \gama~and \cdil~struggle especially in environments with visual observation.
    }
    \label{fig:align_score_p2a}
\end{figure}

\subsection{Alignment Quality}
\label{sec:align}
\paragraph{Quantitative Evaluation}
\label{sec:quantitative}
We evaluate the quality of the alignment to know how well \ours~aligns states in the latent space compared to the baselines, the contribution of each loss to the alignment, and the impact of the alignment on the downstream performance. 
In P2P, as the two domains are essentially identical, we expect that corresponding states are mapped to the same latent representation. 
For evaluation, we sample 1000 corresponding state pairs from domains, encode them into the latent space, 
and test if we can find corresponding states based on the distance in the latent space.
We compare \ours~with the ablated variants of \ours~and the baselines on the top-1 accuracy.
In the other environments, we sample 2000 states and calculate the probability of corresponding states having the top-5\% closeness among all pairs in the latent space. 
Although the detailed cross-domain correspondence is not well-defined in P2A and V2V, we define the correspondence as the xy position in the maze and the end-effector position of the robots, respectively.

The results in Figure \ref{fig:align_score}, \ref{fig:align_score_p2a} show that the regularization terms of \ours~enhances the quality of the alignment, while the contribution of TCC is task-dependent and sometimes TCC disrupts the alignment brought by MMD (in V2V-Open).
Notably, we observe that the alignment of \ours~is significantly enhanced by the BC term. 
Moreover, in the environments that have no action shift or minor action difference between domains, 
\ours~only with the BC term shows as good alignment capability as \ours~or \ourst~with the regularization terms do. 
This result indicates that multi-domain behavioral cloning implicitly contributes to cross-domain representation alignment.
Despite the simpler architecture, \ours, and sometimes even the BC-only variant also, shows comparable or better performance than the existing methods.
The adversarial training of domain translation in V2V environments is unstable and limits the performance of \gama~and \cdil. 
\ours-disc also shows lower performance than \ours~in P2A and V2V, where the domain discrepancy is large 
because the discriminative distribution matching is too restrictive and can hinder appropriate alignment.

\paragraph{Qualitative Evaluation}
\label{sec:qualitative}
We visualize the latent state distributions for Maze experiments in Figure \ref{fig:repr}.
In P2P, the MMD loss effectively aligns the representations.
In P2A, where the domains differ significantly due to varying agent morphologies, 
neither MMD loss nor TCC alone makes a difference we can visually confirm, and we need both to get them closer.
The MMD loss moves distributions toward the same space while preserving their structure.
In contrast, with the discriminative loss, distributions overlap but seem to lose the original structure in each domain, 
resulting in the performance drop in Figure \ref{fig:align_score_p2a} and Table \ref{tbl:success_rates} in the next section. 
In Appendix \ref{app:corresponding_state}, we present interpretable cross-domain correspondence on the plots of R2R and V2V-Reach.

\subsection{Cross-Domain Transfer Performance}
\label{sec:transfer_performance}
\begin{table}[tbp]
\caption{The success rates averaged over nine runs (mean $\pm$ std). 
In P2A-OOD, no method completes the task.
}
\label{tbl:success_rates}
\centering
\begin{tabular}{lccccccc}
\hline
\multicolumn{1}{c}{Task}  & \ours~(Ours) & \ourst~(Ours) &  \gama & \cdil & \cond & BC \\ 
\hline
P2P-umaze  & $0.48\pm0.36$ & $0.68 \pm 0.39$    & $0.71\pm 0.40$ & $\bf1.00\pm 0.00$  & $0.47\pm 0.08$ & $0.28\pm 0.23$ \\
P2P-medium & $\bf0.93\pm0.05$ & $0.84\pm 0.08$  & $0.42\pm 0.19$ & $0.62\pm 0.35$ & $0.28\pm 0.20$ & $0.24\pm 0.22$  \\
P2P-OOD    & $0.42\pm0.47$ & $\bf0.60\pm 0.46$  & $0.31\pm 0.38$ & $0.23\pm 0.37$ & $0.00\pm 0.00$ & $0.00\pm 0.00$ \\
P2A-umaze  & $\bf0.53\pm0.42$ & $0.50\pm 0.41$  & $0.01\pm 0.03$ & $0.01\pm 0.02$ & $0.32\pm 0.25$ & $0.33\pm 0.43$ \\
P2A-medium & $0.54\pm0.26$ & $\bf0.70\pm 0.17$  & $0.00\pm 0.01$ & $0.00\pm 0.00$ & $0.00\pm 0.00$ & $0.17\pm 0.12$ \\ \hline
R2R-Lift   & $0.63\pm0.37$ & $\bf0.71 \pm 0.21$ & $0.09\pm 0.15$ & $0.00\pm 0.00$ & $0.09\pm 0.19$ & $0.22\pm 0.40$ \\
V2V-Reach  & $\bf0.64\pm0.13$ & $0.58\pm 0.14$  & $0.12\pm 0.18$ & $0.03 \pm 0.05$ & $0.14\pm 0.12$ & $0.08\pm 0.06$ \\
V2V-Open   & $\bf0.64\pm0.10$ & $0.33 \pm 0.24$ & $0.07 \pm 0.01$ & $0.00 \pm 0.00$ & $ 0.00 \pm 0.01$ & $0.00 \pm 0.00$ \\
\hline
\end{tabular}
\end{table}

Table \ref{tbl:success_rates} summarizes the success rate of a target task in each setting.
\ours~outperforms the baselines in most settings ranging from cross-morphology transfer to cross-viewpoint transfer.
\gama~and \cdil~show very limited performance in P2A and manipulation tasks including visual-input environments, where cross-domain translation that these methods rely on is not trivial.
\ours~instead reduces each MDP to the common one and avoids this issue. \ours~also utilizes signals from joint multi-domain imitation and does not fully rely on unstable adversarial training.
Moreover, \ours~trains alignment jointly with the action prediction in an end-to-end way rather than chaining domain translation and other components, which is beneficial for error accumulation and better control performance.
\cond~policy struggles to adapt to OOD tasks including V2V because \cond~could does not have a way to adapt to unseen tasks as it needs data from both domains to update the policy. 
We observe that the lack of adaptation phase also negatively affects the precision of control in general,
leading to the suboptimal performance in P2A and R2R.
BC shows transferring capability to some extent but the performance is suboptimal because it updates all parameters at adaptation even though it can enjoy implicit alignment via imitation as \ours.
It also completely fails at highly OOD tasks as it learns OOD target tasks only on the latent space of a source domain when distributions of domains are separate.
We provide the visualization of agents in the tasks in \ref{app:traj_vis}.

\begin{table}[tbp]
\caption{The success rate comparison between ablated variants of \ours~(mean $\pm$ std). 
}
\label{tbl:term_ablation_full}
\centering
\begin{tabular}{lccccccc}
\hline
\multicolumn{1}{c}{Task}     & \ours & \ourst & \ours-disc & \ourst-disc & BC + TCC & BC Only \\
\hline
P2P-obs-medium & $\bf0.94\pm 0.05$ & $0.92\pm 0.09$ & $0.91\pm 0.08$& $0.90\pm 0.12$  & $0.87\pm 0.10$ & $0.92\pm 0.06$ \\
P2P-medium     & $\bf0.93\pm 0.05$ & $0.84\pm 0.08$ & $0.81\pm 0.13$& $0.88\pm 0.09$  & $0.72\pm 0.29$ & $0.48\pm 0.23$  \\
P2A-medium     & $0.54\pm 0.26$ & $\bf0.70\pm 0.17$ & $0.45\pm 0.21$& $0.52\pm 0.15$  & $0.62\pm 0.24$ & $0.52\pm 0.20$  \\ \hline
R2R-Lift       & $0.63\pm 0.37$ & $\bf0.71\pm 0.21$ & $0.11\pm 0.17$& $0.39\pm 0.40$  & $0.54\pm 0.42$ & $0.64\pm 0.35$    \\
V2V-Reach      & $0.64\pm 0.13$ & $0.58\pm 0.14$ & $0.46\pm 0.34$& $\bf0.67\pm 0.18$  & $0.52\pm 0.29$ & $0.50\pm 0.11$ \\
V2V-Open       & $\bf0.64\pm 0.10$ & $0.33\pm 0.24$ & $0.10\pm 0.12$& $0.43\pm 0.14$  & $0.29\pm 0.19$ & $0.02\pm 0.02$   \\
\hline
\\
\end{tabular}
\end{table}

\subsection{Effect of Each Loss Term on Transfer Performance}
To investigate the impact of each loss term on transfer performance, 
we remove terms from the objective or replace the MMD loss with the domain discriminative loss.
Table \ref{tbl:term_ablation_full} presents the success rates. 
As observed in the alignment score evaluation in Section \ref{sec:align}, the MMD effectively enhances the performance of \ours, whereas the contribution of TCC is task-dependent.
The performance drop of \ours-disc in environments with large cross-domain gaps roughly corresponds to the performance drop in the alignment score due to excessively strong regularization. 
The performance gap between the BC baseline in Table \ref{tbl:success_rates} and BC-only showcases the advantage of the hierarchical architecture and training strategy of \ours, 
highlighting the importance of having a fixed shared representation space for cross-domain transfer.
Note that sometimes the alignment score is not directly reflected in the downstream performance 
because the model can project essentially the same states for the action prediction to close positions in the process of abstraction, 
and also the oracle correspondence we assume is not necessarily the best one for the models.

In environments with consistent action format and less-OOD target tasks such as P2P-obs and R2R, BC-only performs similarly to \ours-full as implied by the high alignment scores in Figure \ref{fig:align_score} and \ref{fig:align_score_p2a}.
In V2V environments, the performance of BC-only decreases as the target task is getting OOD, showing the necessity of encouraging cross-domain alignment for such OOD target tasks.
This good performance aligns with the recent success in large-scale imitation across multiple domains and robots with a shared architecture and action format \citep{jang2022bc,Ebert-RSS-22,brohan2022rt}, and extended versions of them \citep{padalkar2023open, team2024octo}.
For additional discussions on the scaling of a dataset, the number of domains and proxy tasks, hyperparameter sensitivity,  
and learning from state-only demonstrations, please refer to Appendix \ref{app:results}.

\section{Conclusion}
In this study, we introduce \ours, a novel method for learning a domain-shared policy for cross-domain policy transfer. 
\ours~leverages implicit representation alignment of multi-domain BC as a main ingredient of cross-domain transfer and supports it with regularization terms such as the MMD loss and TCC.
Our experimental results show the effectiveness of \ours~across situations such as cross-robot, cross-viewpoint, and OOD-task settings. 
Despite its simplicity, \ours~enjoys similar or better cross-domain alignment than existing methods as well as higher precision action prediction thanks to the end-to-end training strategy.
We also confirm that the MMD loss helps to align the latent distributions while keeping their original structure, whereas domain discriminative training can disrupt them when it forces complete overlap.

Although \ours~shows promising results, it has several limitations.
The performance of \ours~is not stable and it fails at adaptation to complex OOD tasks in drastically different domains such as P2A-OOD.
Besides, similar to existing methods, \ours~cannot handle novel groups of states that appear only in the target task, such as new objects. 
Combining a large-scale dataset or pretrained foundation policies with \ours~would be a promising future direction.
To broaden the applicability of \ours, future work could also explore approaches to utilize state-only demonstrations.
In the context of lifelong learning, it is also important to have strategies for continually accepting new domains.
Thanks to its simplicity, we can easily extend \ours~to build our idea on top of it.
We hope that our work provides valuable insights for researchers aiming to develop domain-free portable policies, abstract policies that can be applied to any domain in a zero-shot manner.

\subsubsection*{Acknowledgments}
This work was supported by JSPS KAKENHI Grant Number JP23H03450.

\bibliography{main}

\begin{thebibliography}{60}
\providecommand{\natexlab}[1]{#1}
\providecommand{\url}[1]{\texttt{#1}}
\expandafter\ifx\csname urlstyle\endcsname\relax
  \providecommand{\doi}[1]{doi: #1}\else
  \providecommand{\doi}{doi: \begingroup \urlstyle{rm}\Url}\fi

\bibitem[Andrychowicz et~al.(2020)Andrychowicz, Baker, Chociej, Jozefowicz, McGrew, Pachocki, Petron, Plappert, Powell, Ray, et~al.]{andrychowicz2020learning}
OpenAI:~Marcin Andrychowicz, Bowen Baker, Maciek Chociej, Rafal Jozefowicz, Bob McGrew, Jakub Pachocki, Arthur Petron, Matthias Plappert, Glenn Powell, Alex Ray, et~al.
\newblock Learning dexterous in-hand manipulation.
\newblock \emph{The International Journal of Robotics Research}, 39\penalty0 (1):\penalty0 3--20, 2020.

\bibitem[Arjovsky et~al.(2017)Arjovsky, Chintala, and Bottou]{arjovsky2017wasserstein}
Martin Arjovsky, Soumith Chintala, and L{\'e}on Bottou.
\newblock Wasserstein generative adversarial networks.
\newblock In \emph{International conference on machine learning}, pp.\  214--223. PMLR, 2017.

\bibitem[Baktashmotlagh et~al.(2016)Baktashmotlagh, Harandi, and Salzmann]{baktashmotlagh2016distribution}
Mahsa Baktashmotlagh, Mehrtash Harandi, and Mathieu Salzmann.
\newblock Distribution-matching embedding for visual domain adaptation.
\newblock \emph{Journal of Machine Learning Research}, 17:\penalty0 Article--number, 2016.

\bibitem[Barreto et~al.(2017)Barreto, Dabney, Munos, Hunt, Schaul, van Hasselt, and Silver]{barreto2017successor}
Andr{\'e} Barreto, Will Dabney, R{\'e}mi Munos, Jonathan~J Hunt, Tom Schaul, Hado~P van Hasselt, and David Silver.
\newblock Successor features for transfer in reinforcement learning.
\newblock \emph{Advances in neural information processing systems}, 30, 2017.

\bibitem[Brohan et~al.(2022)Brohan, Brown, Carbajal, Chebotar, Dabis, Finn, Gopalakrishnan, Hausman, Herzog, Hsu, et~al.]{brohan2022rt}
Anthony Brohan, Noah Brown, Justice Carbajal, Yevgen Chebotar, Joseph Dabis, Chelsea Finn, Keerthana Gopalakrishnan, Karol Hausman, Alex Herzog, Jasmine Hsu, et~al.
\newblock {RT-1}: Robotics transformer for real-world control at scale.
\newblock \emph{arXiv preprint arXiv:2212.06817}, 2022.

\bibitem[Castro \& Precup(2010)Castro and Precup]{castro2010using}
Pablo~Samuel Castro and Doina Precup.
\newblock Using bisimulation for policy transfer in mdps.
\newblock In \emph{Twenty-Fourth AAAI Conference on Artificial Intelligence}, 2010.

\bibitem[Courty et~al.(2016)Courty, Flamary, Tuia, and Rakotomamonjy]{courty2016optimal}
Nicolas Courty, R{\'e}mi Flamary, Devis Tuia, and Alain Rakotomamonjy.
\newblock Optimal transport for domain adaptation.
\newblock \emph{IEEE transactions on pattern analysis and machine intelligence}, 39\penalty0 (9):\penalty0 1853--1865, 2016.

\bibitem[Dwibedi et~al.(2019)Dwibedi, Aytar, Tompson, Sermanet, and Zisserman]{dwibedi2019temporal}
Debidatta Dwibedi, Yusuf Aytar, Jonathan Tompson, Pierre Sermanet, and Andrew Zisserman.
\newblock Temporal cycle-consistency learning.
\newblock In \emph{Proceedings of the IEEE/CVF conference on computer vision and pattern recognition}, pp.\  1801--1810, 2019.

\bibitem[Dziugaite et~al.(2015)Dziugaite, Roy, and Ghahramani]{dziugaite2015training}
Gintare~Karolina Dziugaite, Daniel~M. Roy, and Zoubin Ghahramani.
\newblock Training generative neural networks via maximum mean discrepancy optimization.
\newblock In \emph{Proceedings of the Thirty-First Conference on Uncertainty in Artificial Intelligence}, UAI'15, pp.\  258–267, Arlington, Virginia, USA, 2015. AUAI Press.
\newblock ISBN 9780996643108.

\bibitem[Ebert et~al.(2022)Ebert, Yang, Schmeckpeper, Bucher, Georgakis, Daniilidis, Finn, and Levine]{Ebert-RSS-22}
Frederik Ebert, Yanlai Yang, Karl Schmeckpeper, Bernadette Bucher, Georgios Georgakis, Kostas Daniilidis, Chelsea Finn, and Sergey Levine.
\newblock {Bridge Data: Boosting Generalization of Robotic Skills with Cross-Domain Datasets}.
\newblock In \emph{Proceedings of Robotics: Science and Systems}, New York City, NY, USA, June 2022.
\newblock \doi{10.15607/RSS.2022.XVIII.063}.

\bibitem[Fickinger et~al.(2022)Fickinger, Cohen, Russell, and Amos]{fickinger2022crossdomain}
Arnaud Fickinger, Samuel Cohen, Stuart Russell, and Brandon Amos.
\newblock Cross-domain imitation learning via optimal transport.
\newblock In \emph{International Conference on Learning Representations}, 2022.

\bibitem[Franzmeyer et~al.(2022)Franzmeyer, Torr, and Henriques]{franzmeyer2022learn}
Tim Franzmeyer, Philip Torr, and Joao~F. Henriques.
\newblock Learn what matters: cross-domain imitation learning with task-relevant embeddings.
\newblock In Alice~H. Oh, Alekh Agarwal, Danielle Belgrave, and Kyunghyun Cho (eds.), \emph{Advances in Neural Information Processing Systems}, 2022.

\bibitem[Fu et~al.(2020)Fu, Kumar, Nachum, Tucker, and Levine]{fu2020d4rl}
Justin Fu, Aviral Kumar, Ofir Nachum, George Tucker, and Sergey Levine.
\newblock {D4RL}: Datasets for deep data-driven reinforcement learning, 2020.

\bibitem[Ganin et~al.(2016)Ganin, Ustinova, Ajakan, Germain, Larochelle, Laviolette, Marchand, and Lempitsky]{ganin2016domain}
Yaroslav Ganin, Evgeniya Ustinova, Hana Ajakan, Pascal Germain, Hugo Larochelle, Fran{\c{c}}ois Laviolette, Mario Marchand, and Victor Lempitsky.
\newblock Domain-adversarial training of neural networks.
\newblock \emph{The journal of machine learning research}, 17\penalty0 (1):\penalty0 2096--2030, 2016.

\bibitem[Gelada et~al.(2019)Gelada, Kumar, Buckman, Nachum, and Bellemare]{gelada2019deepmdp}
Carles Gelada, Saurabh Kumar, Jacob Buckman, Ofir Nachum, and Marc~G Bellemare.
\newblock {DeepMDP}: Learning continuous latent space models for representation learning.
\newblock In \emph{International Conference on Machine Learning}, pp.\  2170--2179. PMLR, 2019.

\bibitem[Gretton et~al.(2012)Gretton, Borgwardt, Rasch, Sch{\"o}lkopf, and Smola]{gretton2012kernel}
Arthur Gretton, Karsten~M Borgwardt, Malte~J Rasch, Bernhard Sch{\"o}lkopf, and Alexander Smola.
\newblock A kernel two-sample test.
\newblock \emph{The Journal of Machine Learning Research}, 13\penalty0 (1):\penalty0 723--773, 2012.

\bibitem[Gupta et~al.(2017)Gupta, Devin, Liu, Abbeel, and Levine]{gupta2017learning}
Abhishek Gupta, Coline Devin, YuXuan Liu, Pieter Abbeel, and Sergey Levine.
\newblock Learning invariant feature spaces to transfer skills with reinforcement learning.
\newblock In \emph{International Conference on Learning Representations}, 2017.

\bibitem[Gupta et~al.(2022)Gupta, Fan, Ganguli, and Fei-Fei]{gupta2022metamorph}
Agrim Gupta, Linxi Fan, Surya Ganguli, and Li~Fei-Fei.
\newblock {MetaMorph}: Learning universal controllers with transformers.
\newblock In \emph{International Conference on Learning Representations}, 2022.

\bibitem[Hejna et~al.(2020)Hejna, Pinto, and Abbeel]{hejna2020hierarchically}
Donald Hejna, Lerrel Pinto, and Pieter Abbeel.
\newblock Hierarchically decoupled imitation for morphological transfer.
\newblock In \emph{International Conference on Machine Learning}, pp.\  4159--4171. PMLR, 2020.

\bibitem[Hendrycks \& Gimpel(2016)Hendrycks and Gimpel]{hendrycks2016gaussian}
Dan Hendrycks and Kevin Gimpel.
\newblock Gaussian error linear units (gelus).
\newblock \emph{arXiv preprint arXiv:1606.08415}, 2016.

\bibitem[Ho \& Ermon(2016)Ho and Ermon]{ho2016generative}
Jonathan Ho and Stefano Ermon.
\newblock Generative adversarial imitation learning.
\newblock \emph{Advances in neural information processing systems}, 29, 2016.

\bibitem[Hoffman et~al.(2018)Hoffman, Tzeng, Park, Zhu, Isola, Saenko, Efros, and Darrell]{hoffman2018cycada}
Judy Hoffman, Eric Tzeng, Taesung Park, Jun-Yan Zhu, Phillip Isola, Kate Saenko, Alexei Efros, and Trevor Darrell.
\newblock {CyCADA}: Cycle-consistent adversarial domain adaptation.
\newblock In \emph{International conference on machine learning}, pp.\  1989--1998. Pmlr, 2018.

\bibitem[Jang et~al.(2021)Jang, Irpan, Khansari, Kappler, Ebert, Lynch, Levine, and Finn]{jang2022bc}
Eric Jang, Alex Irpan, Mohi Khansari, Daniel Kappler, Frederik Ebert, Corey Lynch, Sergey Levine, and Chelsea Finn.
\newblock {BC-Z}: Zero-shot task generalization with robotic imitation learning.
\newblock In \emph{Conference on Robot Learning}, pp.\  991--1002. PMLR, 2021.

\bibitem[Kim et~al.(2020)Kim, Gu, Song, Zhao, and Ermon]{kim2020domain}
Kuno Kim, Yihong Gu, Jiaming Song, Shengjia Zhao, and Stefano Ermon.
\newblock Domain adaptive imitation learning.
\newblock In \emph{International Conference on Machine Learning}, pp.\  5286--5295. PMLR, 2020.

\bibitem[Kumar et~al.(2019)Kumar, Fu, Soh, Tucker, and Levine]{kumar2019stabilizing}
Aviral Kumar, Justin Fu, Matthew Soh, George Tucker, and Sergey Levine.
\newblock Stabilizing off-policy q-learning via bootstrapping error reduction.
\newblock \emph{Advances in Neural Information Processing Systems}, 32, 2019.

\bibitem[Li et~al.(2015)Li, Swersky, and Zemel]{li2015generative}
Yujia Li, Kevin Swersky, and Rich Zemel.
\newblock Generative moment matching networks.
\newblock In \emph{International conference on machine learning}, pp.\  1718--1727. PMLR, 2015.

\bibitem[Liu et~al.(2020)Liu, Ling, Mu, and Su]{Liu2020State}
Fangchen Liu, Zhan Ling, Tongzhou Mu, and Hao Su.
\newblock State alignment-based imitation learning.
\newblock In \emph{International Conference on Learning Representations}, 2020.

\bibitem[Liu et~al.(2018{\natexlab{a}})Liu, Lehman, Molino, Petroski~Such, Frank, Sergeev, and Yosinski]{liu2018intriguing}
Rosanne Liu, Joel Lehman, Piero Molino, Felipe Petroski~Such, Eric Frank, Alex Sergeev, and Jason Yosinski.
\newblock An intriguing failing of convolutional neural networks and the coordconv solution.
\newblock \emph{Advances in neural information processing systems}, 31, 2018{\natexlab{a}}.

\bibitem[Liu et~al.(2017)Liu, Bousquet, and Chaudhuri]{liu2017approximation}
Shuang Liu, Olivier Bousquet, and Kamalika Chaudhuri.
\newblock Approximation and convergence properties of generative adversarial learning.
\newblock \emph{Advances in Neural Information Processing Systems}, 30, 2017.

\bibitem[Liu et~al.(2018{\natexlab{b}})Liu, Gupta, Abbeel, and Levine]{liu2018imitation}
YuXuan Liu, Abhishek Gupta, Pieter Abbeel, and Sergey Levine.
\newblock Imitation from observation: Learning to imitate behaviors from raw video via context translation.
\newblock In \emph{2018 IEEE International Conference on Robotics and Automation (ICRA)}, pp.\  1118--1125. IEEE, 2018{\natexlab{b}}.

\bibitem[Long et~al.(2013)Long, Wang, Ding, Sun, and Yu]{long2013transfer}
Mingsheng Long, Jianmin Wang, Guiguang Ding, Jiaguang Sun, and Philip~S Yu.
\newblock Transfer feature learning with joint distribution adaptation.
\newblock In \emph{Proceedings of the IEEE international conference on computer vision}, pp.\  2200--2207, 2013.

\bibitem[Long et~al.(2015)Long, Cao, Wang, and Jordan]{long2015learning}
Mingsheng Long, Yue Cao, Jianmin Wang, and Michael Jordan.
\newblock Learning transferable features with deep adaptation networks.
\newblock In \emph{International conference on machine learning}, pp.\  97--105. PMLR, 2015.

\bibitem[Loshchilov \& Hutter(2019)Loshchilov and Hutter]{loshchilov2018decoupled}
Ilya Loshchilov and Frank Hutter.
\newblock Decoupled weight decay regularization.
\newblock In \emph{International Conference on Learning Representations}, 2019.
\newblock URL \url{https://openreview.net/forum?id=Bkg6RiCqY7}.

\bibitem[{Octo Model Team}(2024)]{team2024octo}
{Octo Model Team}.
\newblock Octo: An open-source generalist robot policy.
\newblock \emph{arXiv preprint arXiv:2405.12213}, 2024.

\bibitem[Padalkar et~al.(2023)Padalkar, Pooley, Jain, Bewley, Herzog, Irpan, Khazatsky, Rai, Singh, Brohan, et~al.]{padalkar2023open}
Abhishek Padalkar, Acorn Pooley, Ajinkya Jain, Alex Bewley, Alex Herzog, Alex Irpan, Alexander Khazatsky, Anant Rai, Anikait Singh, Anthony Brohan, et~al.
\newblock {Open X-Embodiment}: Robotic learning datasets and {RT-X} models.
\newblock \emph{arXiv preprint arXiv:2310.08864}, 2023.

\bibitem[Peng et~al.(2018)Peng, Andrychowicz, Zaremba, and Abbeel]{peng2018sim}
Xue~Bin Peng, Marcin Andrychowicz, Wojciech Zaremba, and Pieter Abbeel.
\newblock Sim-to-real transfer of robotic control with dynamics randomization.
\newblock In \emph{2018 IEEE international conference on robotics and automation (ICRA)}, pp.\  3803--3810. IEEE, 2018.

\bibitem[Raffin et~al.(2021)Raffin, Hill, Gleave, Kanervisto, Ernestus, and Dormann]{stable-baselines3}
Antonin Raffin, Ashley Hill, Adam Gleave, Anssi Kanervisto, Maximilian Ernestus, and Noah Dormann.
\newblock {Stable-Baselines3}: Reliable reinforcement learning implementations.
\newblock \emph{Journal of Machine Learning Research}, 22\penalty0 (268):\penalty0 1--8, 2021.
\newblock URL \url{http://jmlr.org/papers/v22/20-1364.html}.

\bibitem[Rao et~al.(2020)Rao, Harris, Irpan, Levine, Ibarz, and Khansari]{rao2020rl}
Kanishka Rao, Chris Harris, Alex Irpan, Sergey Levine, Julian Ibarz, and Mohi Khansari.
\newblock {RL-CycleGAN}: Reinforcement learning aware simulation-to-real.
\newblock In \emph{Proceedings of the IEEE/CVF Conference on Computer Vision and Pattern Recognition}, pp.\  11157--11166, 2020.

\bibitem[Raychaudhuri et~al.(2021)Raychaudhuri, Paul, Vanbaar, and Roy-Chowdhury]{raychaudhuri2021cross}
Dripta~S Raychaudhuri, Sujoy Paul, Jeroen Vanbaar, and Amit~K Roy-Chowdhury.
\newblock Cross-domain imitation from observations.
\newblock In \emph{International Conference on Machine Learning}, pp.\  8902--8912. PMLR, 2021.

\bibitem[Schulman et~al.(2017)Schulman, Wolski, Dhariwal, Radford, and Klimov]{schulman2017proximal}
John Schulman, Filip Wolski, Prafulla Dhariwal, Alec Radford, and Oleg Klimov.
\newblock Proximal policy optimization algorithms.
\newblock \emph{arXiv preprint arXiv:1707.06347}, 2017.

\bibitem[Sermanet et~al.(2018)Sermanet, Lynch, Chebotar, Hsu, Jang, Schaal, Levine, and Brain]{sermanet2018time}
Pierre Sermanet, Corey Lynch, Yevgen Chebotar, Jasmine Hsu, Eric Jang, Stefan Schaal, Sergey Levine, and Google Brain.
\newblock Time-contrastive networks: Self-supervised learning from video.
\newblock In \emph{2018 IEEE international conference on robotics and automation (ICRA)}, pp.\  1134--1141. IEEE, 2018.

\bibitem[Stadie et~al.(2017)Stadie, Abbeel, and Sutskever]{stadie2017third}
Bradly~C Stadie, Pieter Abbeel, and Ilya Sutskever.
\newblock Third person imitation learning.
\newblock In \emph{International Conference on Learning Representations}, 2017.

\bibitem[Sun \& Saenko(2016)Sun and Saenko]{sun2016deep}
Baochen Sun and Kate Saenko.
\newblock Deep {CORAL}: Correlation alignment for deep domain adaptation.
\newblock In \emph{Computer Vision--ECCV 2016 Workshops: Amsterdam, The Netherlands, October 8-10 and 15-16, 2016, Proceedings, Part III 14}, pp.\  443--450. Springer, 2016.

\bibitem[Sun et~al.(2016)Sun, Feng, and Saenko]{sun2016return}
Baochen Sun, Jiashi Feng, and Kate Saenko.
\newblock Return of frustratingly easy domain adaptation.
\newblock In \emph{Proceedings of the AAAI conference on artificial intelligence}, volume~30, 2016.

\bibitem[Sun et~al.(2022)Sun, Zheng, Wang, Cohen, and Huang]{sun2022transfer}
Yanchao Sun, Ruijie Zheng, Xiyao Wang, Andrew~E Cohen, and Furong Huang.
\newblock Transfer {RL} across observation feature spaces via model-based regularization.
\newblock In \emph{International Conference on Learning Representations}, 2022.

\bibitem[Tobin et~al.(2017)Tobin, Fong, Ray, Schneider, Zaremba, and Abbeel]{tobin2017domain}
Josh Tobin, Rachel Fong, Alex Ray, Jonas Schneider, Wojciech Zaremba, and Pieter Abbeel.
\newblock Domain randomization for transferring deep neural networks from simulation to the real world.
\newblock In \emph{2017 IEEE/RSJ international conference on intelligent robots and systems (IROS)}, pp.\  23--30. IEEE, 2017.

\bibitem[Tzeng et~al.(2014)Tzeng, Hoffman, Zhang, Saenko, and Darrell]{tzeng2014deep}
Eric Tzeng, Judy Hoffman, Ning Zhang, Kate Saenko, and Trevor Darrell.
\newblock Deep domain confusion: Maximizing for domain invariance.
\newblock \emph{arXiv preprint arXiv:1412.3474}, 2014.

\bibitem[Tzeng et~al.(2017)Tzeng, Hoffman, Saenko, and Darrell]{tzeng2017adversarial}
Eric Tzeng, Judy Hoffman, Kate Saenko, and Trevor Darrell.
\newblock Adversarial discriminative domain adaptation.
\newblock In \emph{Proceedings of the IEEE conference on computer vision and pattern recognition}, pp.\  7167--7176, 2017.

\bibitem[Van~der Maaten \& Hinton(2008)Van~der Maaten and Hinton]{van2008visualizing}
Laurens Van~der Maaten and Geoffrey Hinton.
\newblock Visualizing data using {t-SNE}.
\newblock \emph{Journal of machine learning research}, 9\penalty0 (11), 2008.

\bibitem[Vaswani et~al.(2017)Vaswani, Shazeer, Parmar, Uszkoreit, Jones, Gomez, Kaiser, and Polosukhin]{vaswani2017attention}
Ashish Vaswani, Noam Shazeer, Niki Parmar, Jakob Uszkoreit, Llion Jones, Aidan~N Gomez, {\L}ukasz Kaiser, and Illia Polosukhin.
\newblock Attention is all you need.
\newblock \emph{Advances in neural information processing systems}, 30, 2017.

\bibitem[Wang et~al.(2019)Wang, Sun, and Halgamuge]{wang2018improving}
Wei Wang, Yuan Sun, and Saman Halgamuge.
\newblock Improving {MMD}-{GAN} training with repulsive loss function.
\newblock In \emph{International Conference on Learning Representations}, 2019.
\newblock URL \url{https://openreview.net/forum?id=HygjqjR9Km}.

\bibitem[Wang et~al.(2021)Wang, Li, Ding, Nie, Chen, Dong, and Wang]{wang2021rethinking}
Wei Wang, Haojie Li, Zhengming Ding, Feiping Nie, Junyang Chen, Xiao Dong, and Zhihui Wang.
\newblock Rethinking maximum mean discrepancy for visual domain adaptation.
\newblock \emph{IEEE Transactions on Neural Networks and Learning Systems}, 34\penalty0 (1):\penalty0 264--277, 2021.

\bibitem[Wang et~al.(2022)Wang, Cao, Hao, and Sadigh]{wang2022weakly}
Zihan Wang, Zhangjie Cao, Yilun Hao, and Dorsa Sadigh.
\newblock Weakly supervised correspondence learning.
\newblock In \emph{2022 International Conference on Robotics and Automation (ICRA)}, pp.\  469--476, 2022.
\newblock \doi{10.1109/ICRA46639.2022.9811729}.

\bibitem[Yin et~al.(2022)Yin, Sun, Ma, Tomizuka, and Li]{yin2022cross}
Zhao-Heng Yin, Lingfeng Sun, Hengbo Ma, Masayoshi Tomizuka, and Wu-Jun Li.
\newblock Cross domain robot imitation with invariant representation.
\newblock In \emph{2022 International Conference on Robotics and Automation (ICRA)}, pp.\  455--461. IEEE, 2022.

\bibitem[Yu et~al.(2019)Yu, Quillen, He, Julian, Hausman, Finn, and Levine]{yu2020meta}
Tianhe Yu, Deirdre Quillen, Zhanpeng He, Ryan Julian, Karol Hausman, Chelsea Finn, and Sergey Levine.
\newblock {Meta-World}: A benchmark and evaluation for multi-task and meta reinforcement learning.
\newblock In \emph{Conference on robot learning}, pp.\  1094--1100. PMLR, 2019.

\bibitem[Zakka et~al.(2021)Zakka, Zeng, Florence, Tompson, Bohg, and Dwibedi]{zakka2022xirl}
Kevin Zakka, Andy Zeng, Pete Florence, Jonathan Tompson, Jeannette Bohg, and Debidatta Dwibedi.
\newblock {XIRL}: Cross-embodiment inverse reinforcement learning.
\newblock In \emph{Conference on Robot Learning}, pp.\  537--546. PMLR, 2021.

\bibitem[Zhang et~al.(2020)Zhang, Lyle, Sodhani, Filos, Kwiatkowska, Pineau, Gal, and Precup]{zhang2020invariant}
Amy Zhang, Clare Lyle, Shagun Sodhani, Angelos Filos, Marta Kwiatkowska, Joelle Pineau, Yarin Gal, and Doina Precup.
\newblock Invariant causal prediction for block mdps.
\newblock In \emph{International Conference on Machine Learning}, pp.\  11214--11224. PMLR, 2020.

\bibitem[Zhang et~al.(2021)Zhang, Xiao, Efros, Pinto, and Wang]{zhang2021learning}
Qiang Zhang, Tete Xiao, Alexei~A Efros, Lerrel Pinto, and Xiaolong Wang.
\newblock Learning cross-domain correspondence for control with dynamics cycle-consistency.
\newblock In \emph{International Conference on Learning Representations}, 2021.

\bibitem[Zhu et~al.(2017)Zhu, Park, Isola, and Efros]{Zhu_2017_ICCV}
Jun-Yan Zhu, Taesung Park, Phillip Isola, and Alexei~A. Efros.
\newblock Unpaired image-to-image translation using cycle-consistent adversarial networks.
\newblock In \emph{Proceedings of the IEEE International Conference on Computer Vision (ICCV)}, Oct 2017.

\bibitem[Zhu et~al.(2020)Zhu, Wong, Mandlekar, Mart\'{i}n-Mart\'{i}n, Joshi, Nasiriany, and Zhu]{robosuite2020}
Yuke Zhu, Josiah Wong, Ajay Mandlekar, Roberto Mart\'{i}n-Mart\'{i}n, Abhishek Joshi, Soroush Nasiriany, and Yifeng Zhu.
\newblock robosuite: A modular simulation framework and benchmark for robot learning.
\newblock In \emph{arXiv preprint arXiv:2009.12293}, 2020.

\end{thebibliography}
\bibliographystyle{collas2024_conference}

\newpage
\appendix
\section{Additional Algorithm Table}
\label{app:algorithm}

\begin{algorithm}[h]
\caption{Task Adaptation with Imitation}
\label{alg:adaptation}
\begin{algorithmic}
  \REQUIRE State encoder $\phi(s_d, d)$, Common policy $\pi_z(s_z, k)$, Action decoder $\psi(a_z, s_d, d)$ trained by Alg. \ref{alg:alignment}.
  \REQUIRE Trajectories of target task $k'$: $\mathcal{D}_{x,k'} = \{(s_x, a_x, x, k')\}$.
  [Optional] Dataset $\mathcal{D}$ used in Alg. \ref{alg:alignment} to regulate $\pi_z$.
  \STATE Freeze parameters of state encoder $\phi$ and action decoder $\psi$.
  \WHILE{not converged}
    \STATE Sample a batch of state action pairs. $\{(s_x, a_x, x, k')\} \sim \mathcal{D}_{x,k'}$
    \STATE [Optional] Sample a batch from $\mathcal{D}$ used for alignment and add it to the batch from $\mathcal{D}_{x,k'}$.
    \STATE Calculate $\mathcal{L}_\text{BC}$ in (\ref{eq:adapt}) on the batch.
    \STATE Update parameters of $\pi_z(s_z, k)$ only.
  \ENDWHILE
\end{algorithmic}
\end{algorithm}

\section{Additional Illustrations, Results and Discussions}
\label{app:results}

\subsection{Alignment Score with Top-5 Accuracy}
\begin{figure}[h]
    \centering
    \includegraphics[width=.85\textwidth]{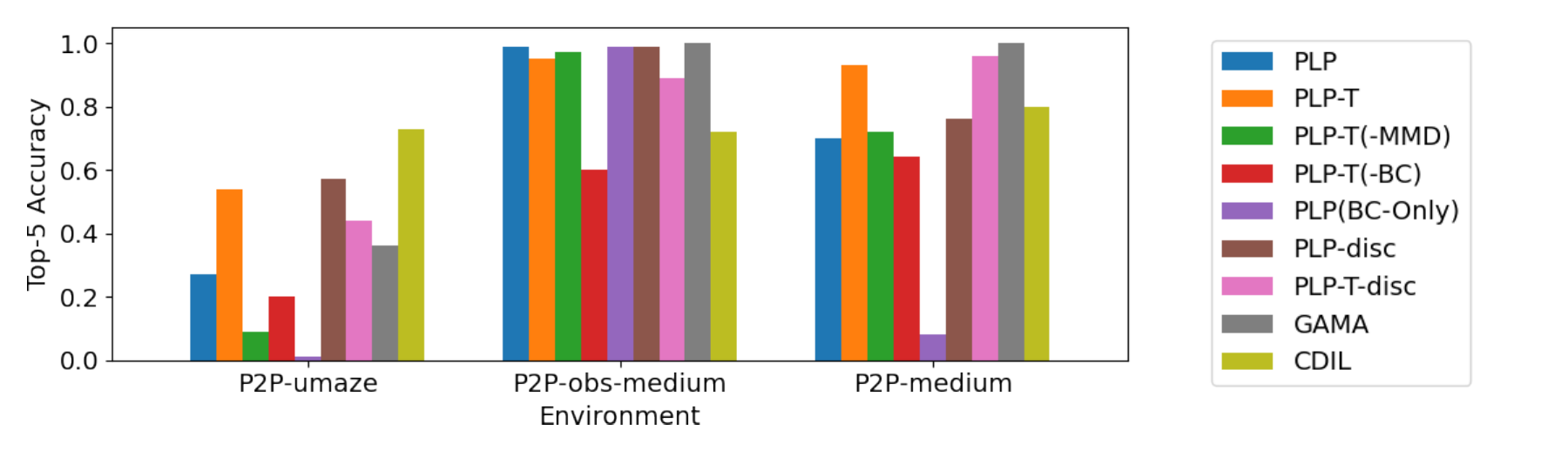}
    \caption{Alignment scores. The values are top-5 accuracies of finding a corresponding state from 1k random states based on the distance in the latent space (or state space for \gama).}
    \label{fig:top5}
\end{figure}

Figure \ref{fig:top5} shows the alignment scores when we use top-5 accuracy instead of top-1 accuracy as a metric.
The results show the same tendency as the one with top-1 accuracy (c.f. Figure \ref{fig:align_score}).

\subsection{Additional Maze Experiments with Spider robot.}
We test the transfer performance from an agent with four legs (Ant) to an agent with six legs (Spider). 
We call this task Ant-to-Spider (A2S).
The robots are morphologically similar, but the positions of the legs are different from each other.
Table \ref{tbl:spider} shows the results. \ours-performs best among the methods.

\begin{table}[h]
\caption{The success rates of the Ant-to-Spider (A2S) task averaged over nine runs. }
\label{tbl:spider}
\centering
\begin{tabular}{lccccccc}
\hline
\multicolumn{1}{c}{Task}     & \ourst~(Ours)  & \gama & \cdil & \cond & BC \\ 
\hline
A2S-umaze & $\bf0.80\pm 0.30$ & $0.10\pm 0.14$ & $0.00\pm 0.00$  & $0.01\pm 0.01$ & $0.42\pm 0.38$ \\
A2S-medium & $\bf0.45\pm 0.18$ & $0.04\pm 0.05$ & $0.00\pm 0.00$ & $0.00\pm 0.00$ & $0.25\pm 0.20$  \\
\hline
\end{tabular}
\end{table}

\subsection{Alignment Complexity}
\label{app:align_complexity}
\begin{figure}[tb]
    \begin{minipage}{0.49\textwidth}
        \centering
        \includegraphics[width=.75\textwidth]{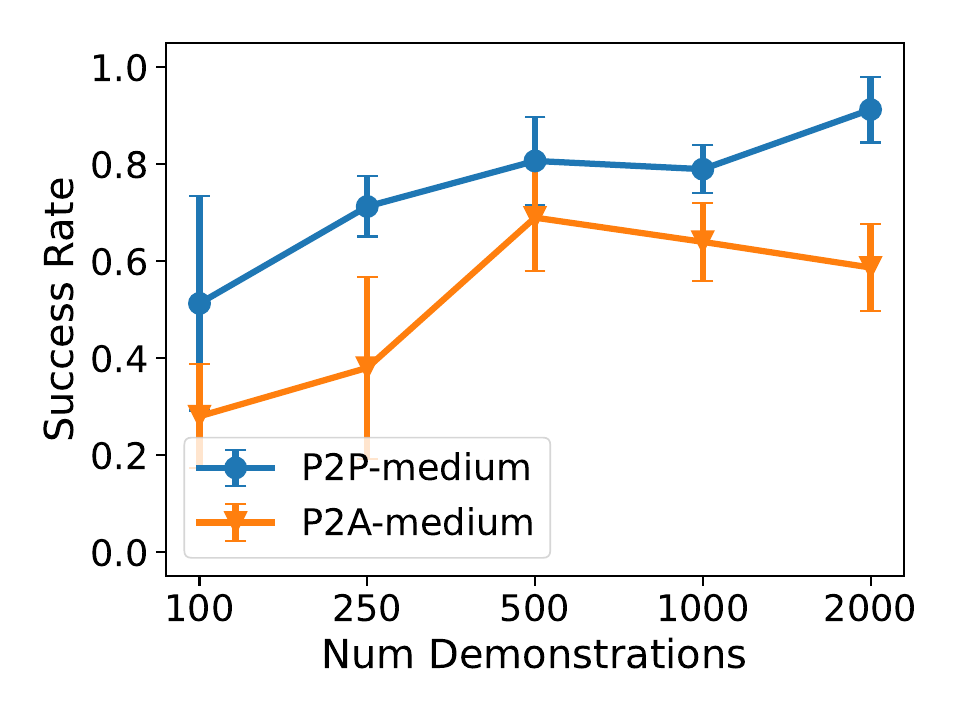}
        \subcaption{The number of demonstrations vs success rate}
        \label{fig:traj_o}
    \end{minipage}
    \begin{minipage}{0.49\textwidth}
        \centering
        \includegraphics[width=.75\textwidth]{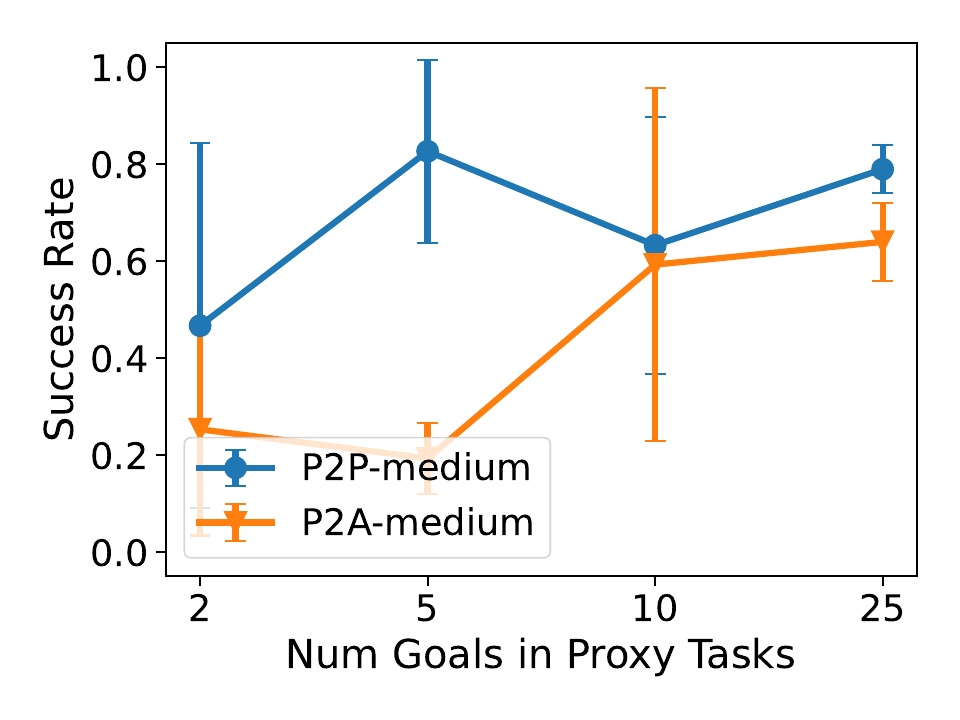}
        \subcaption{The number of goals in contained in proxy tasks vs success rate}
        \label{fig:task_o}
    \end{minipage}
    \caption{Alignment complexity in the medium maze.
     The scores are averaged over three runs with a fixed single goal, 
     and the error bars represent the standard deviations.
    }
    \label{fig:complexity}
\end{figure}

Figure \ref{fig:complexity} shows the alignment complexity in P2P-medium and P2A-medium concerning the number of demonstrations and the number of proxy tasks.
Note that the number of goals in the figure corresponds to one-fourth of the number of proxy tasks.
We can observe that an increase in the number of demonstrations or proxy tasks positively affects performance,
but the performance improvement does not continue until the perfect transfer especially in P2A seemingly due to 
the insufficient quality of alignment.

\begin{figure}[tbp]
    \begin{minipage}{0.49\textwidth}
        \centering
        \includegraphics[width=.75\textwidth]{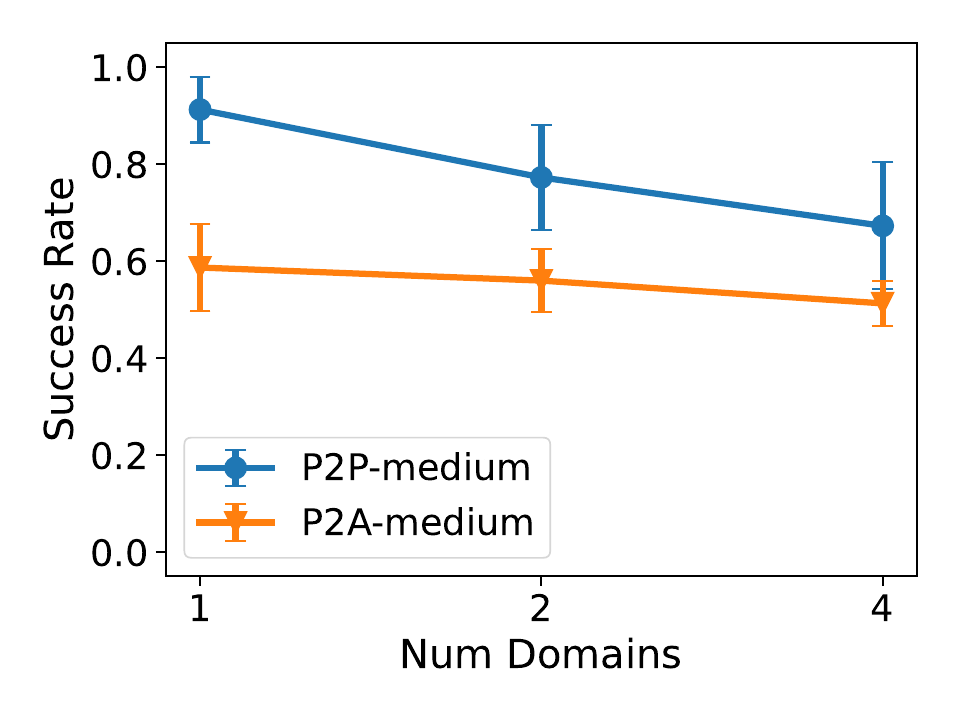}
        \subcaption{The number of domains vs success rate, when the total number of trajectories is kept constant.}
        \label{fig:domain_normal}
    \end{minipage}
    \begin{minipage}{0.49\textwidth}
        \centering
        \includegraphics[width=.75\textwidth]{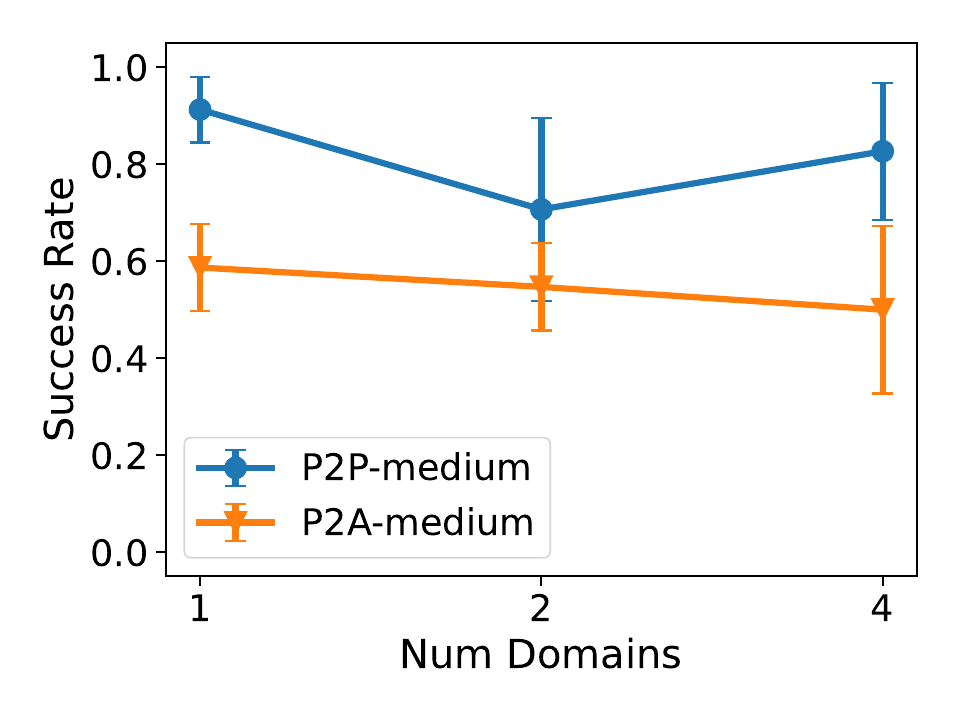}
        \subcaption{The number of domains vs success rate, when the total number of trajectories increases proportionally.}
        \label{fig:domain_large}
    \end{minipage}
    \caption{The relationship between the number of source domains and the performance.
     The scores are averaged over three runs with a fixed single goal.
     The error bars show the standard deviations.
    }
    \label{fig:abl_domain}
\end{figure}

\subsection{Effect of the Number of Source Domains}
One possible advantage of \ours~compared to methods like \gama~is that we can easily increase the number of domains
contained in the source dataset.
However, it is unclear how the increase in domain variations affects imitation performance.
To investigate this, we measure it in two settings: 
i) increase the number of domains in a fixed-size dataset 
ii) increase the number of domains while keeping the number of trajectories from one domain.
In ii), the dataset size increases proportionally to the number of domains. 
For the comparison, we maintain the total number of epochs, as we observe that the training with the larger dataset does not converge within the same number of iterations.

The results measured in P2P-medium and P2A-medium are summarized in Figure \ref{fig:abl_domain}.
The increase in domain variations makes it difficult to shape a good shared representation space, resulting in a performance drop. 
As shown in Figure \ref{fig:domain_large}, if the number of demonstrations for each domain is maintained, the performance degradation is mitigated, 
although it does not improve the performance compared to the single-domain case, either.
Future work could explore methods that can effectively leverage multi-source datasets to boost the performance of a transferred policy.

\begin{figure}[tbp]
    \begin{minipage}{0.49\textwidth}
        \centering
        \includegraphics[width=.75\textwidth]{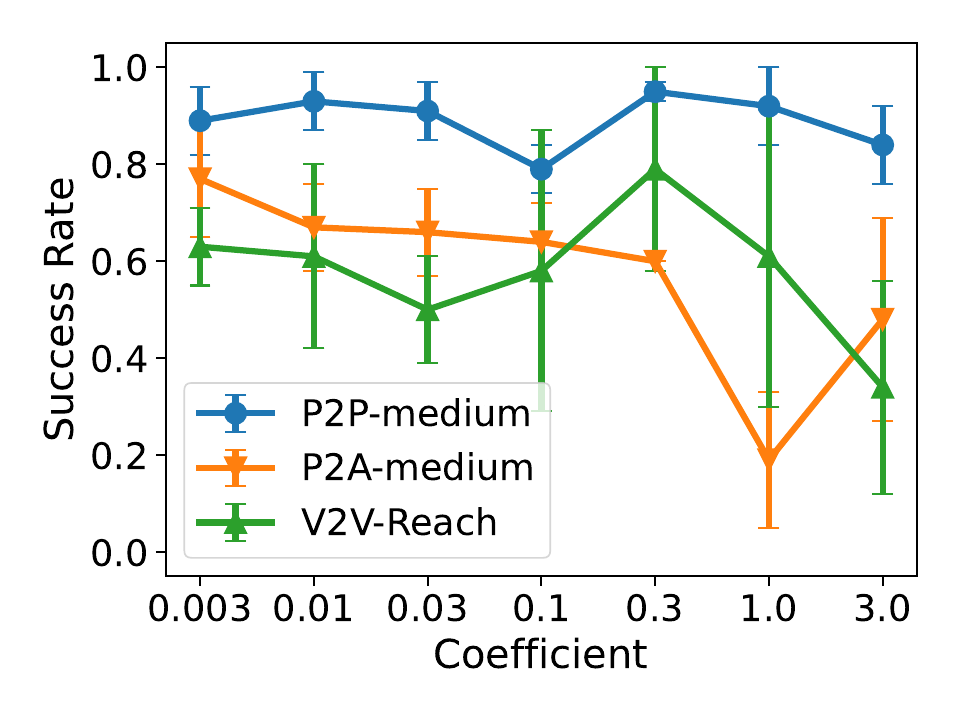}
        \subcaption{The coefficient for MMD $\lambda_\text{MMD}$ vs success rate.}
        \label{fig:mmd_coef}
    \end{minipage}
    \begin{minipage}{0.49\textwidth}
        \centering
        \includegraphics[width=.75\textwidth]{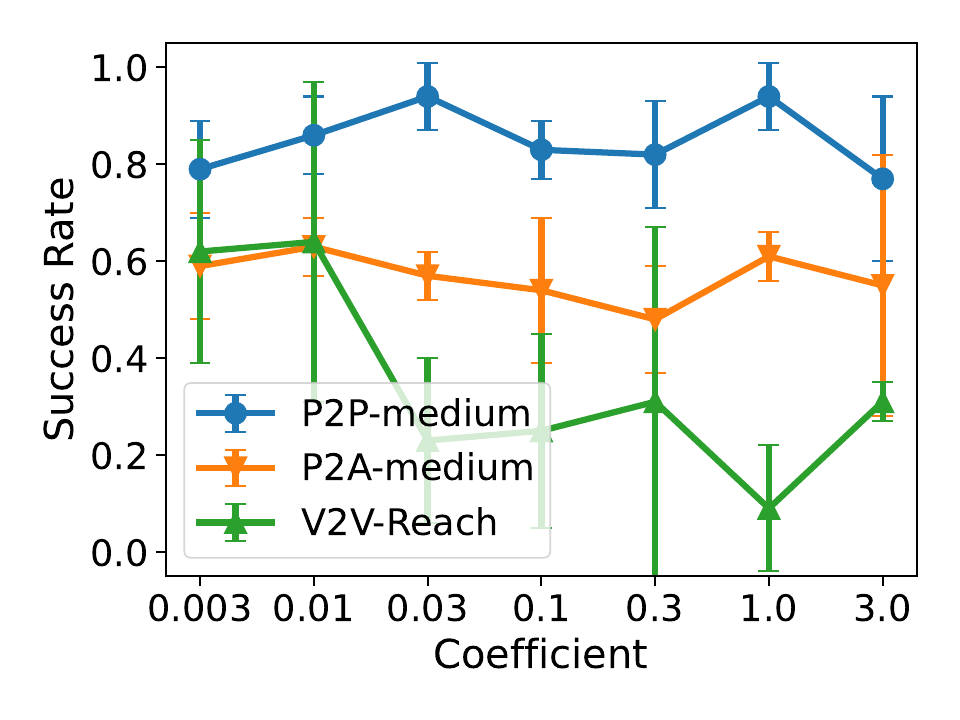}
        \subcaption{The coefficient for the discriminative loss in \ours-disc vs success rate.}
        \label{fig:adv_coef}
    \end{minipage}
    \caption{The relationship between the coefficient of the regularization term and the performance.
     The scores are averaged over three runs with a fixed single goal.
     The error bars show the standard deviations.
    }
    \label{fig:abl_param}
\end{figure}

\subsection{Effect of the Coefficient Hyperparameter}
Figure \ref{fig:abl_param} shows the performance changes when we sweep the coefficient of the regularization term for the MMD of \ours~or the discriminative objective of \ours-disc.
We confirmed that the performance of \ours~is not highly sensitive to the choice of $\lambda_\text{MMD}$
For \ours-disc in V2V-Reach, we observe the performance degradation when we set the coefficient to the large value, so we use 0.01 for the experiments.

\begin{table}[tbp]
\caption{The success rates of \ours~with the state input for the action decoder in R2R and V2V-Reach.}
\label{tbl:with_state}
\centering
\begin{tabular}{lccc}
\hline
\multicolumn{1}{c}{Task} & \ourst  & \ourst~+ State & \ourst~+ State (Proprio. only) \\
\hline
R2R-Lift      & $\bf0.71 \pm 0.21$ & $0.53 \pm 0.48$ & N/A  \\  
V2V-Reach     & $\bf0.68 \pm 0.24$ & $0.38 \pm 0.25$ & $0.64 \pm 0.30$   \\
\hline
\end{tabular}
\end{table}

\subsection{\ours~with State Input for Decoder in R2R-Lift and V2V-Reach
}
As we mentioned in the last part of Section \ref{sec:explanation_alignment}, the inclusion of state input for the decoder can sometimes lead to performance degradation.
This occurs because the decoder can receive all necessary information except the task ID even without the common policy.
While we show the performance without state input in the main results of 
R2R and V2V,
here we present the performance with state input in Table \ref{tbl:with_state}.
We observe a performance drop in environments, although it still significantly outperforms the baselines.
For V2V-Reach, we also test a variant that exclusively receives proprioceptive inputs necessary for precise action prediction in the decoder, omitting the image inputs.
This modification yields improved performance comparable to the full \ours. 
This highlights the potential for performance enhancement in \ours~when we have prior knowledge about which inputs aid action prediction and which elements should be aligned across domains to facilitate better transfer.

\subsection{\gama~with State Input for Decoder}
The reason why \gama~does not perform well in environments like P2A is the lack of symmetry and exact correspondence between domains, 
resulting in the failure of action translation of \gama~from the source domain to the target domain.
In contrast, \ours~overcomes this problem by learning a common latent space and focusing on transferring shared structure between domains.
Another explanation of the performance gap between \ours~and \gama~is that 
\ours~provides domain-specific state information to the decoder while \gama~does not.
To investigate the impact of state input on the performance of \ours, 
we measure the performance of \gama~when it receives state input for the action translation function of \gama.
The results are presented in Table \ref{tbl:gama_s}. 
With the inclusion of the state input, \gama~is able to bridge the complexity gap between the source and target domain to some extent. 
However, it still faces difficulties in finding good correspondences between domains.

\begin{table}[tbp]
\caption{The success rates of \gama~with the state input for the action translation function.
Note that, in R2R-Lift, we do not provide states into the action decoder.}
\label{tbl:gama_s}
\centering
\begin{tabular}{lcccccc}
\hline
\multicolumn{1}{c}{Task} & \ourst  & \gama & \gama~+ State \\
\hline
P2A-umaze      & $\bf0.50 \pm 0.41$ & $0.01 \pm 0.03$ & $0.31 \pm 0.24$  \\  
P2A-medium     & $\bf0.70 \pm 0.17$ & $0.00 \pm 0.01$ & $0.11 \pm 0.07$   \\
R2R-Lift       & $\bf0.71 \pm 0.21$ & $0.09 \pm 0.15$ & $0.21 \pm 0.31$ \\
\hline
\end{tabular}
\end{table}

\subsection{Transfer from State-Only Source Domain}
\begin{table}[tbp]
\caption{Success rate with state-only source domain demonstrations}
\label{tbl:noaction}
\centering
\begin{tabular}{lcccc}
\hline
\multicolumn{1}{c}{Task} & \ourst  & \ourst-noaction \\ 
\hline
P2P-medium     & $0.84 \pm 0.08$ & $0.17 \pm 0.20$ \\
P2A-medium     & $0.70 \pm 0.17$ & $0.14 \pm 0.08$ \\
\hline
\end{tabular}
\end{table}

We can extend the applicability of \ours~to scenarios where actions from the source domains are unavailable.
In such cases, we can replace action prediction with next-state prediction. 
We decouple a \ours~policy into next state prediction and inverse dynamics model.
We calculate the loss on next state prediction in the representation space on the output of a common policy
and optimize the encoder $\phi$ and common policy $\pi_z$.
\begin{align*}
    \mathcal{L}_\text{next\_state} = \mathbb{E}\left[ \|\pi_z(\phi_d(s_d), k) - \text{sg}[\phi_d(s_d')]\|^2 \right],
\end{align*}
where $s_d'$ is a next state of $s_d$ in domain $d$, and $\text{sg}[\cdot]$ shows stopping gradient.
At the same time, we separately train an inverse dynamics model $\psi$ on the target domain dataset optimizing the following objective:
\begin{align*}
\mathcal{L}_\text{IDM} = \mathbb{E}\left[ \| \psi(\text{sg}[\phi_y(s_y)], s_y') -  a_y \|^2 \right]
\end{align*}
At inference, the encoder and common policy predict the next latent state given a current state, domain, and task ID,
followed by the action prediction of the inverse dynamics model from a predicted next latent state and observed current state.

Table \ref{tbl:noaction} presents the results. The performance is limited compared to the original setting.
It is seemingly because this state-only setting misses the opportunity to leverage alignment signal from end-to-end behavioral cloning that the original \ours~takes advantage of.
Additional techniques are required to seamlessly bridge domains with and without action information.
We leave it for future work as mentioned in the Conclusion.

\section{Additional Visualization}
\label{app:visualization}
\subsection{Distribution of Latent Representation and State Correspondence}
\label{app:corresponding_state}

\begin{figure}[tbp]
    \centering
    \includegraphics[width=0.7\textwidth]{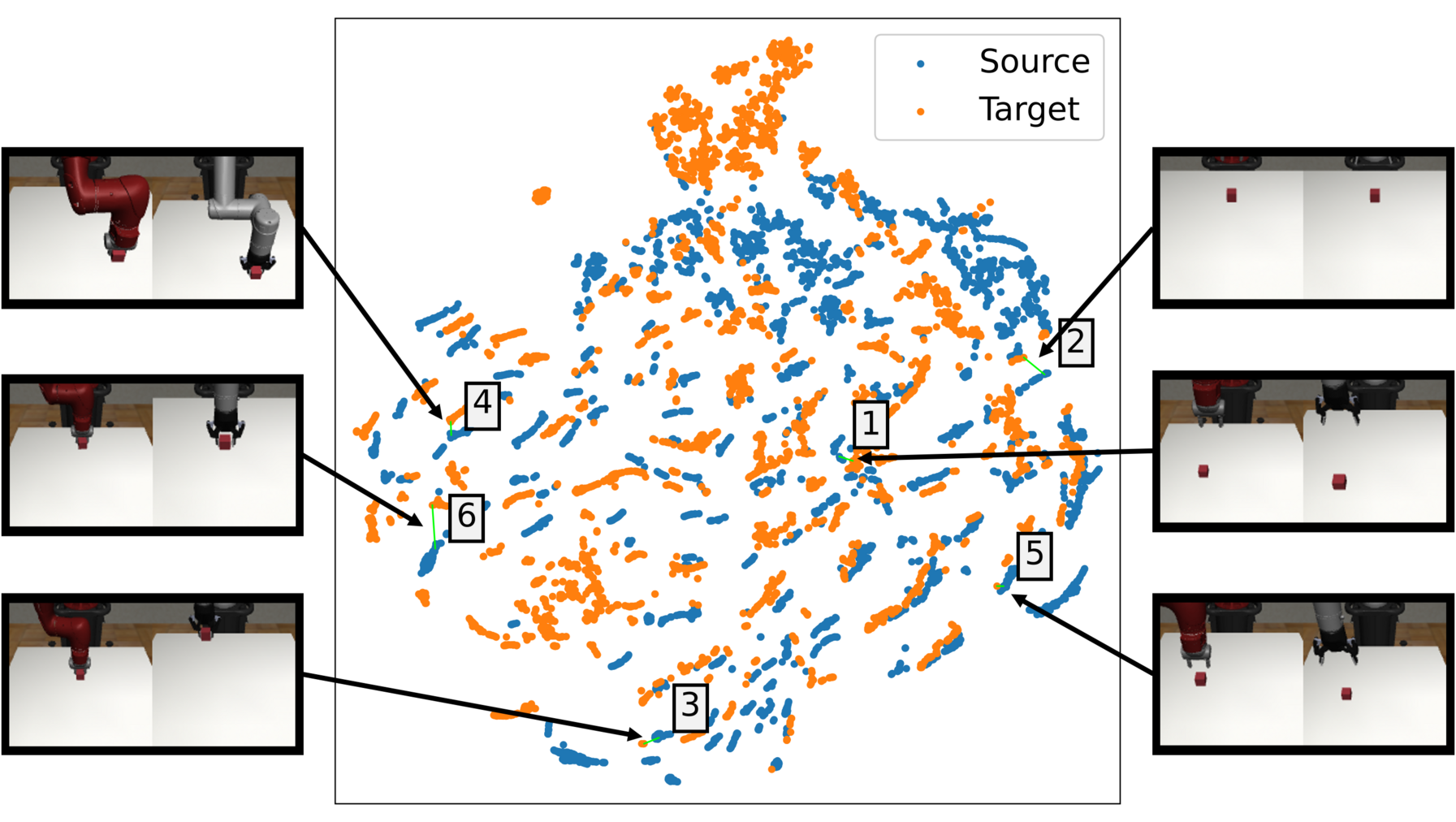}
    \caption{Visualization of the latent space by t-SNE in the R2R-Lift task. 
    We visualize the states of the source (left-hand side) and target (right-hand side) domains that are closest to each other in the latent space. 
    In some cases, the arms were in similar positions across domains (1, 2, 4, 5, 6), 
    but in other cases, we observed positional deviation of the arms (3).}
    \label{fig:r2r_nearest_points}
\end{figure}

In Figure \ref{fig:r2r_nearest_points}, we visualize corresponding states in the latent space of \ours~obtained in R2R-Lift.
With the help of MMD loss, the latent state distributions are overlapped.
In addition, while there are instances of misalignment of states in some areas, we observe cross-domain state-to-state correspondence in arm positions.

\begin{figure}[tb]
    \centering
    \begin{minipage}{0.32\textwidth}
        \centering
        \includegraphics[width=.95\textwidth]{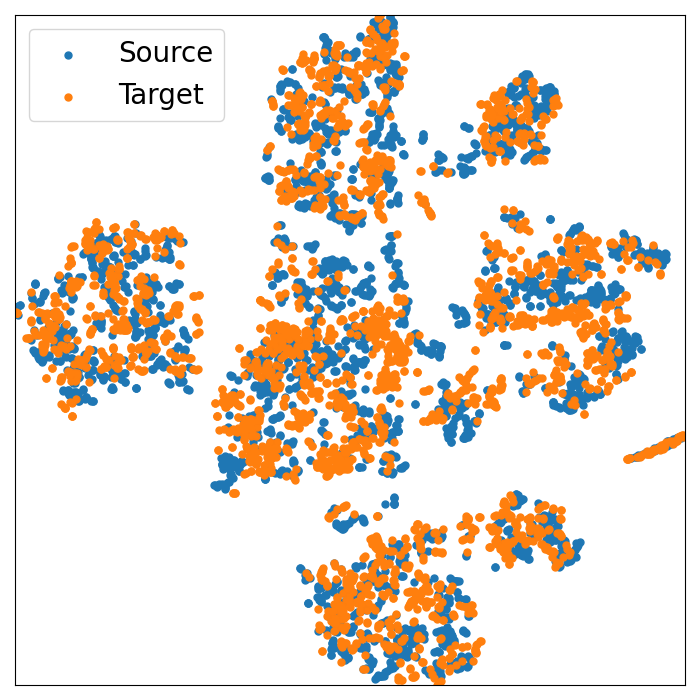}
        \subcaption{All points}
        \label{fig:v2v_color_dist_all}
    \end{minipage}
    \begin{minipage}{0.33\textwidth}
        \centering
        \includegraphics[width=.95\textwidth]{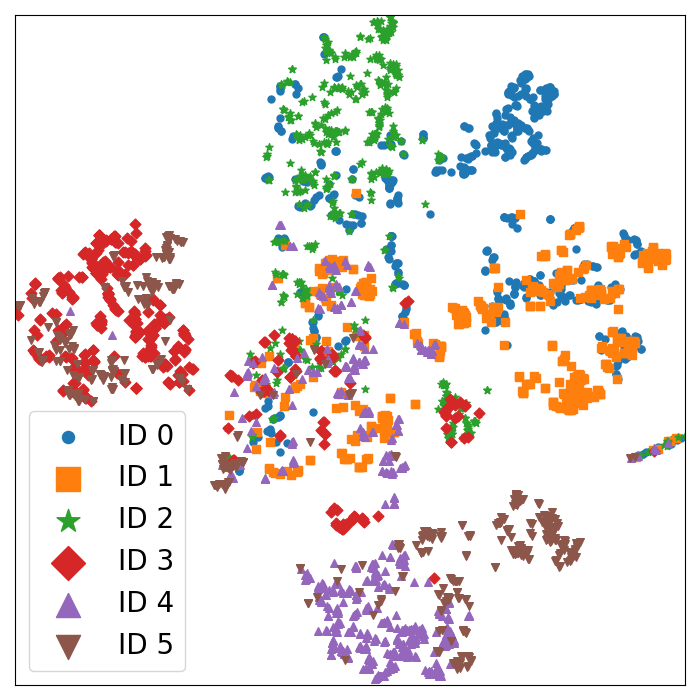}
        \subcaption{Source domain}
        \label{fig:v2v_color_dist_source}
    \end{minipage}
    \begin{minipage}{0.33\textwidth}
        \centering
        \includegraphics[width=.95\textwidth]{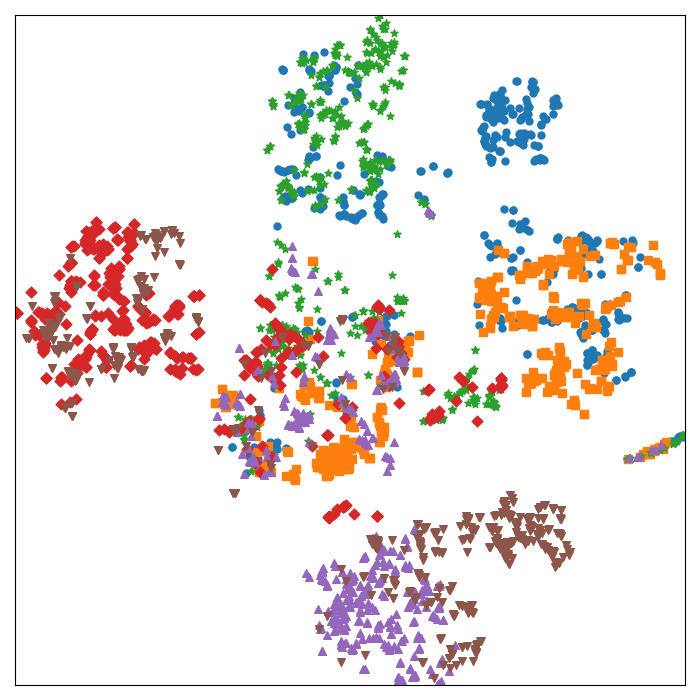}
        \subcaption{Target domain}
        \label{fig:v2v_color_dist_target}
    \end{minipage}
    \caption{Distributions of latent states $s_z$ in V2V-Reach visualized by t-SNE.
    In (\ref{fig:v2v_color_dist_source}) and (\ref{fig:v2v_color_dist_target}), the points are colored according to the color order of the balls in the environment.
    ID 0-5 corresponds to GBR, GRB, BGR, BRG, RGB, and RBG, respectively.
    Each ID roughly forms a cluster in a consistent position across domains.
    }
    \label{fig:latent_vis_v2v_color}
\end{figure}

In Figure \ref{fig:latent_vis_v2v_color}, we visualized the latent state representation of \ours~obtained in V2V-Reach. 
We color representation points for different ball orders with different colors in each domain
how \ours~organize and align the latent space.
We observe that points for each color order form a cluster in the latent space, 
and the positions of clusters for the same arrangement of balls are located in a similar position across domains (\ref{fig:v2v_color_dist_source}, \ref{fig:v2v_color_dist_target}).

\subsection{Visualizations of Trajectories of Agents}
\label{app:traj_vis}
\begin{figure}[tb]
    \centering
    \begin{minipage}{\textwidth}
        \centering
        \includegraphics[width=.9\textwidth]{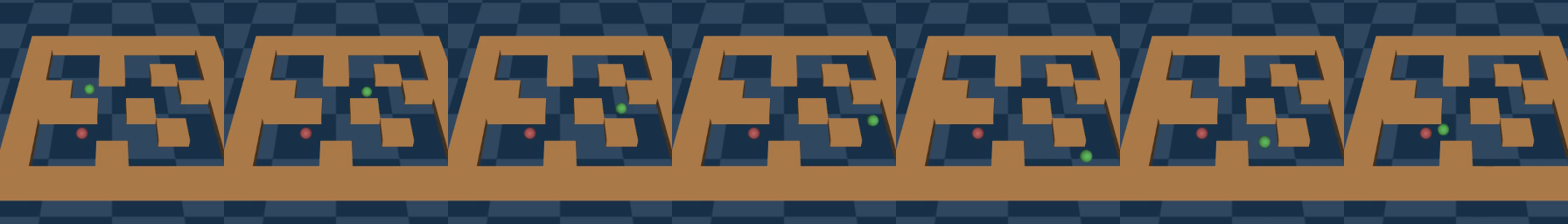}
        \subcaption{A trajectory of \ours}
        \label{fig:traj_p2p_ood_ours}
    \end{minipage}

    \begin{minipage}{\textwidth}
        \centering
        \includegraphics[width=.9\textwidth]{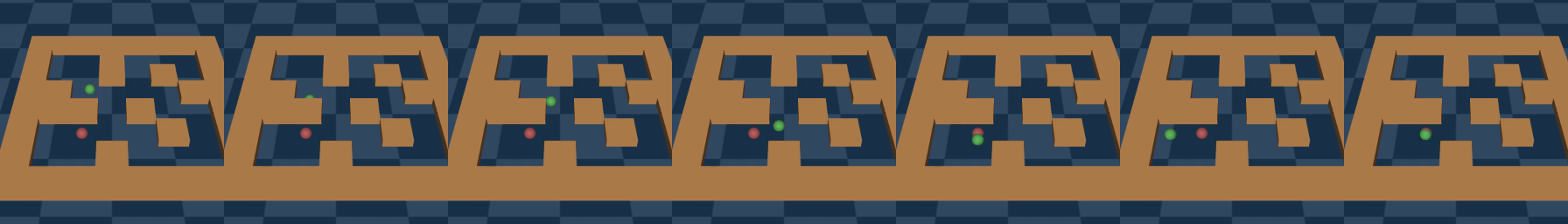}
        \subcaption{A trajectory of \cond}
        \label{fig:traj_p2p_ood_cond}
    \end{minipage}
    \caption{Trajectories of agents in P2P-OOD. The red point shows the goal. The target route is illustrated by the red arrow in Figure \ref{fig:detour}.}
    \label{fig:traj_p2p_ood}
\end{figure}

\begin{figure}[tb]
    \centering
    \begin{minipage}{\textwidth}
        \centering
        \includegraphics[width=.9\textwidth]{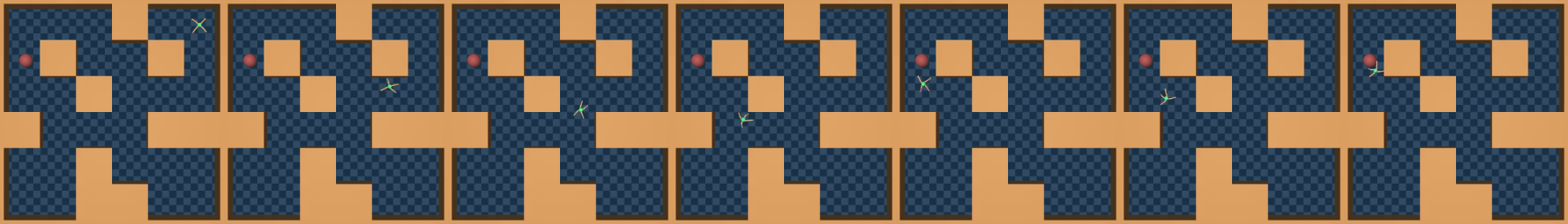}
        \subcaption{A trajectory of \ours}
        \label{fig:traj_p2a_ours}
    \end{minipage}

    \begin{minipage}{\textwidth}
        \centering
        \includegraphics[width=.9\textwidth]{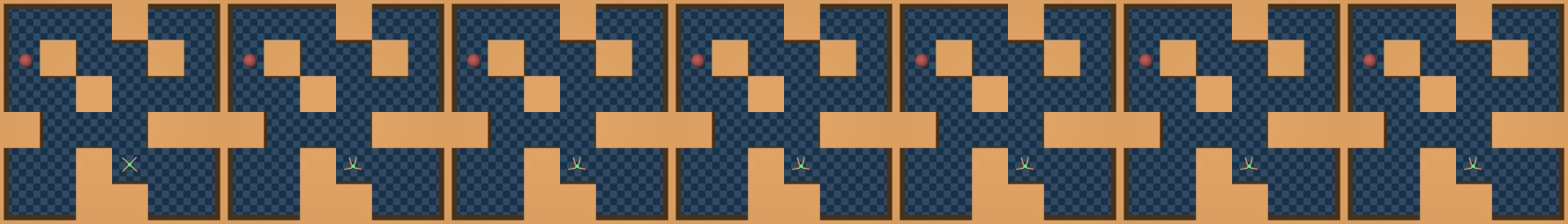}
        \subcaption{A trajectory of \gama}
        \label{fig:traj_p2a_gama}
    \end{minipage}
    \caption{Trajectories of agents in P2A-medium. The red point shows the goal.}
    \label{fig:traj_p2a}
\end{figure}

We visualize the trajectories of the agents to observe their actual behavior.
In Figure \ref{fig:traj_p2p_ood}, we display the trajectories of \ours~and \cond~in P2P-OOD. 
We confirm that \ours~successfully follows the target route, while \cond~fails to adapt to this out-of-distribution target route.
In Figure \ref{fig:traj_p2a}, we present the trajectories of \ours~and \gama~in P2A-medium. 
We observe that the \ours~agent successfully reaches the goal, 
while the \gama~agent cannot move from the starting position due to the challenge of finding exact correspondence between the Point agent and the Ant agent.

\begin{figure}[tb]
    \centering
    \begin{minipage}{0.9\textwidth}
        \centering
        \includegraphics[width=\textwidth]{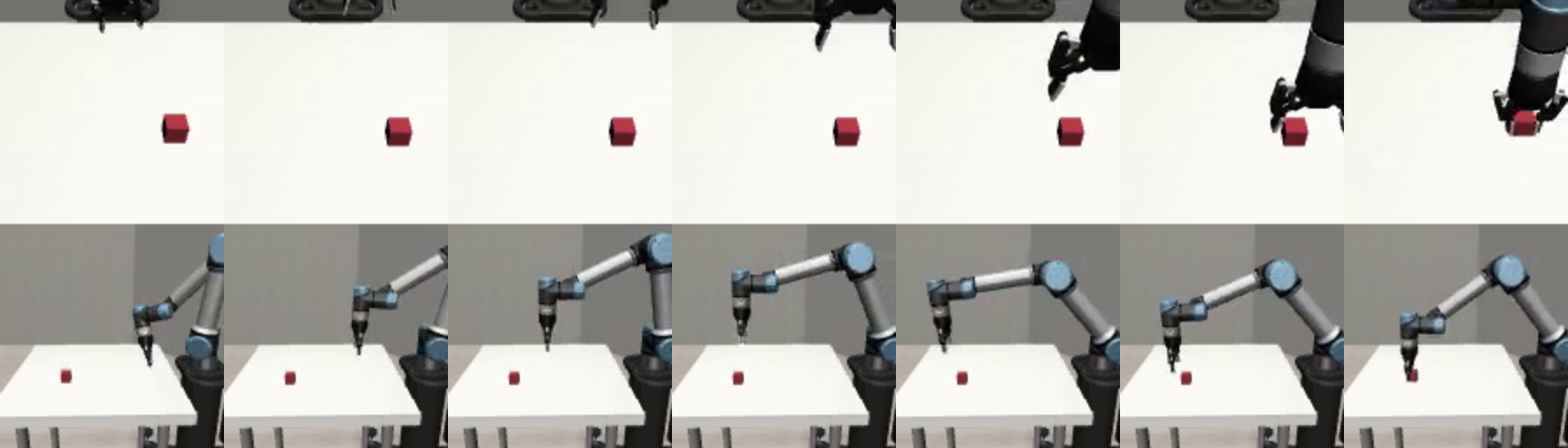}
        \subcaption{A trajectory of \ours}
        \label{fig:traj_r2r_lift_success}
    \end{minipage}

    \begin{minipage}{0.9\textwidth}
        \centering
        \includegraphics[width=\textwidth]{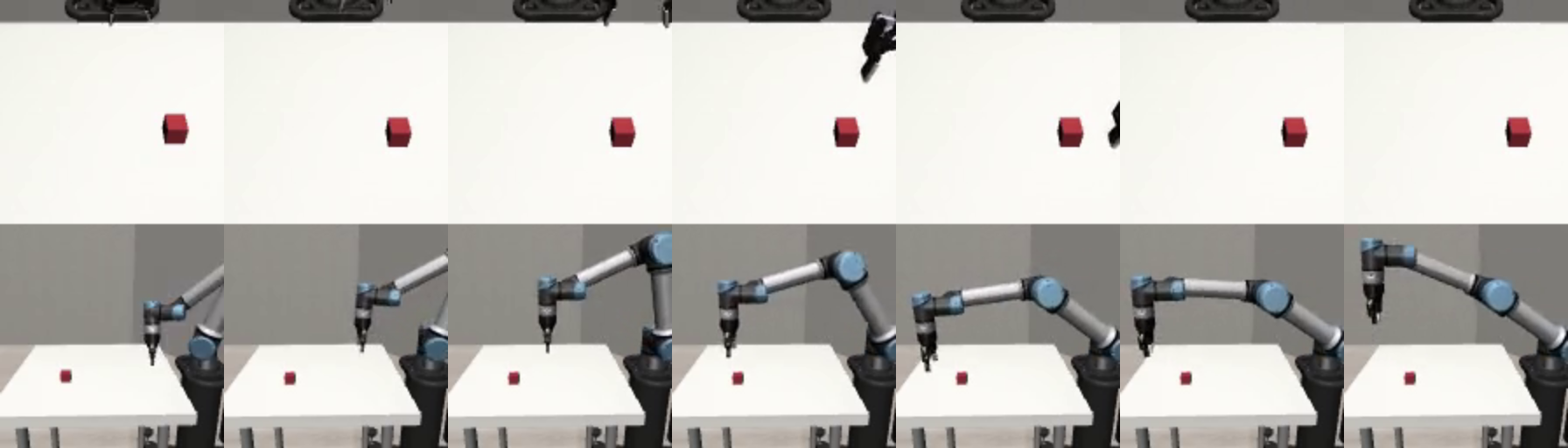}
        \subcaption{A trajectory of \gama}
        \label{fig:traj_r2r_lift_failure}
    \end{minipage}
    \caption{Trajectories of agents in R2R-Lift.}
    \label{fig:traj_r2r_lift}
\end{figure}

Figure \ref{fig:traj_r2r_lift} demonstrates trajectories of \ours~agent and \gama~agent in R2R-Lift. 
\ours~agent precisely operates its arm and successfully grasps the target object, while \gama~agent attempts to close its gripper in the wrong position.

\begin{figure}[tb]
    \centering
    \begin{minipage}{0.9\textwidth}
        \centering
        \includegraphics[width=\textwidth]{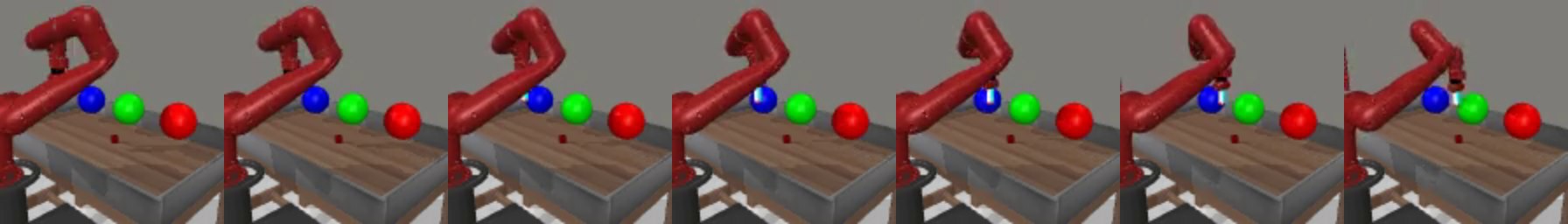}
        \subcaption{A trajectory of \ours}
        \label{fig:traj_color_plp}
    \end{minipage}

    \begin{minipage}{0.9\textwidth}
        \centering
        \includegraphics[width=\textwidth]{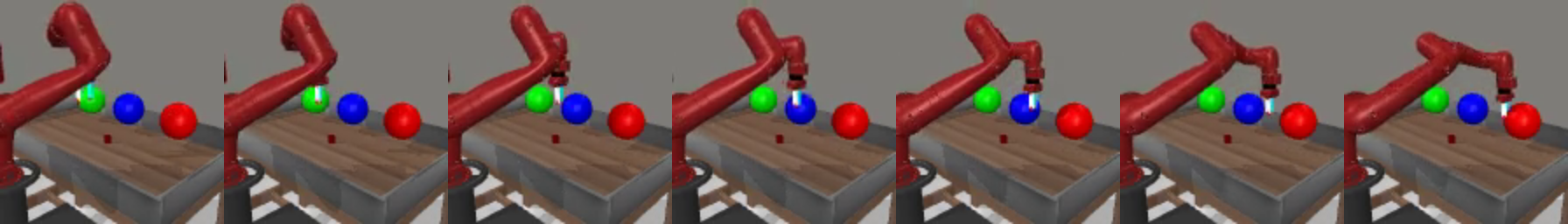}
        \subcaption{A trajectory of BC}
        \label{fig:traj_color_bc}
    \end{minipage}
    \caption{Trajectories of agents in V2V-Reach. The goal color is set to green.}
    \label{fig:traj_color}
\end{figure}

\begin{figure}[tb]
    \centering
    \begin{minipage}{0.9\textwidth}
        \centering
        \includegraphics[width=\textwidth]{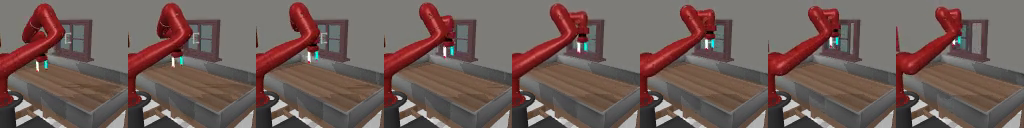}
        \subcaption{V2V-Open Success (\ours)}
        \label{fig:traj_v2v_open_success}
    \end{minipage}

    \begin{minipage}{0.9\textwidth}
        \centering
        \includegraphics[width=\textwidth]{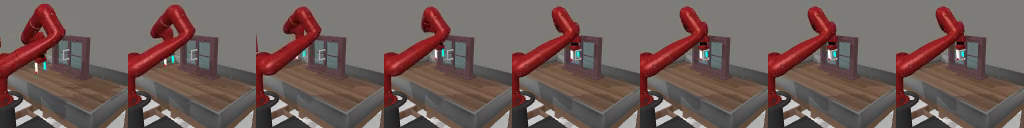}
        \subcaption{V2V-Open Failure (BC)}
        \label{fig:traj_v2v_open_failure_bc}
    \end{minipage}

    \begin{minipage}{0.9\textwidth}
        \centering
        \includegraphics[width=\textwidth]{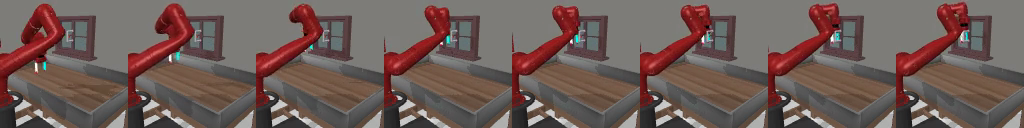}
        \subcaption{V2V-Open Failure (\cond)}
        \label{fig:traj_v2v_open_failure}
    \end{minipage}
    \caption{Trajectories of agents in V2V-Open.}
    \label{fig:traj_v2v_open}
\end{figure}

Figure \ref{fig:traj_color} shows the trajectories of \ours~agent and BC agent in V2V-Reach. \ours~agent successfully adapts to the target task, while BC agent still heads to the goal of a proxy task. 
It is because the adaptation only happens on the source domain states as a BC agent learns about the source domain and target domain separately without aligning representations.
We observe the same tendency in the trajectories of V2V-Open in Figure \ref{fig:traj_v2v_open} as well.
The \ours~agent successfully recognizes the position of the window from the image and accomplishes the task (Figure \ref{fig:traj_v2v_open_success}).
The BC agent and \cond~agent stick to the movement in proxy tasks and move their arm in the opposite direction due to the lack of adaptation capability to OOD movements (\ref{fig:traj_v2v_open_failure_bc}, \ref{fig:traj_v2v_open_failure}).

\section{Experiment Details}
\label{app:experimental_details}
\subsection{Environments}
\label{app:env}
\begin{figure}[tb]
    \centering
    \includegraphics[width=.35\textwidth]{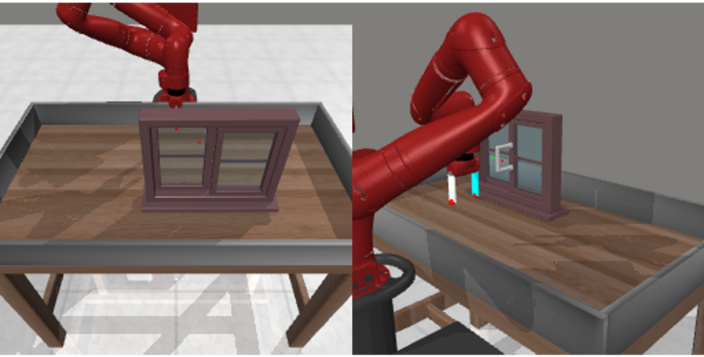}
    \caption{
    Visualization of two different viewpoints in V2V. 
    The agent is required to learn a target task from one viewpoint 
    and perform the task in another domain with a different viewpoint. 
    The figure on the left shows the viewpoint in the source domain, 
    while the figure on the right shows the one in the target domain. 
     }
    \label{fig:metaworld_different}
\end{figure}
All Maze environments used in this paper are based on D4RL \citep{fu2020d4rl}.
These environments involve two types of agents: Point and Ant.
The Point agent has a state space of four dimensions and an action space of two dimensions.
The state space comprises the positions and velocities along the $x$-axis and $y$-axis, 
while the action space consists of the forces to be applied in each direction.
The Ant agent has a state space of 29 dimensions and an action space of eight dimensions.
The state space includes the position and velocity of the body, as well as the joint angles and angular velocities of the four legs. 
The action space consists of the forces applied to each joint.
The task is defined by a combination of a starting area and a specific goal location. 
In the umaze, there are three starting zones and seven goals, while in the medium maze, 
there are four starting zones and 26 goals. 
For the experiments, three goals from different areas in the mazes are selected.
In P2P, we constructed the source domain by swapping the $x$ and $y$-axis of observations and multiplying each element of actions by -1. 
For example, a state-action pair in the target domain $((x, y, v_x, v_y), (a_x, a_y))$ corresponds to $((y, x, v_y, v_x), (-a_x, -a_y))$ in the source domain. 
Here, $x$ and $y$ represent coordinate values instead of domains.
In the OOD variants, agents are required to take a detour in the medium maze as depicted in Figure \ref{fig:detour}, 
instead of directly heading toward the goal via the shortest path.
In P2A, although the shapes of the mazes are consistent between Point and Ant, the scale and x, y directions of the mazes are different between agents from the beginning.

For R2R-Lift, we use robosuite framework~\citep{robosuite2020}.
The Sawyer robot and the UR5e robot are used for the source and target domain, respectively, to test cross-robot transfer (Figure \ref{fig:r2r}).
The observation spaces of these robots consist of the positions or angular information, as well as the velocity of each joint and the end effector.
Sawyer and UR5e have state spaces of 32 dimensions and 37 dimensions, respectively.
Both robots are controlled by delta values of the 3D position of the end effector and a 1D gripper state.
We choose the Block Lifting task, where the robot needs to pick up a block and lift it to a certain height.
A task is defined by the initial position of the object to be lifted.
We set up a total of 27 tasks by placing the blocks on a $3\times 3\times 3$ grid of points on the table.
One position is selected for the target task, and the remaining points that are not at the same height or the same 2D position as the selected position are used as the proxy tasks.
The initial pose of the robot is randomized.

For V2V-Reach and V2V-Open, we use environments from the Meta-World benchmark~\citep{yu2020meta}.
Different viewpoints are employed for the source and target domain as depicted in Figure \ref{fig:metaworld_different}.
The robot observes the proprioceptive state and an image from a specific viewpoint.
A proprioceptive state of the robot consists of a 3D position of the end effector and a 1D gripper state observed in the last two steps.
The action space of the robot is 4-dimensional, representing the delta values of the 3D position of the end effector and the 1D gripper state.
In V2V-Reach, three balls of different colors spawn in a random order. The task is to reach a ball of a specific color. The agent needs to interpret visual observation correctly.
The initial position of the end-effector is also randomized.

In V2V-Open, we use Window-Close for the proxy task and Window-Open for the target task.
When closing the window, the robot approaches the left-hand side of a door and moves toward the right, whereas when opening the window, it heads to the right-hand side and then moves to the left. 
That is, the robot has to take out-of-distribution actions in the target task.
Specifically, Window-Close is divided into four proxy tasks by roughly grouping the position of the window to facilitate representation alignment. 
The success rate of a proxy task (i.e., one area) without visual input is approximately 30\%.
In V2V-Open, which is a target task, we sample the window position from the entire area randomly.

In the evaluation, we measure the success rates with 100 trials in each seed, and we show the average and standard deviations calculated over nine seeds.

\subsection{Datasets}
As explained in Section \ref{sec:formulation}, the dataset comprises state-action trajectories of expert demonstrations encompassing multiple tasks with different goals. 
Specifically, for the P2P and P2A, we provided about 1k trajectories from 18 tasks (six training goals and three starting zones) for the umaze, and about 1k trajectories from 100 tasks (25 training goals and four starting zones) for the medium maze unless stated otherwise in the ablation study. 
The expert demonstrations of the Point agent were downloaded from \url{http://rail.eecs.berkeley.edu/datasets/offline_rl/maze2d/} (maze2d-umaze-sparse-v1, maze2d-medium-sparse-v1). 
When generating expert trajectories for the Ant agent, we employed PPO \citep{schulman2017proximal} from stable-baselines3 \citep{stable-baselines3} to train agents to move one of the four cardinal directions (up, down, left, or right).
Subsequently, we constructed complete demonstrations by solving the maze using breadth-first search (BFS) and providing the agent with the direction of the next square.
For R2R and V2V tasks, we collected expert trajectories with scripted policies based on the object position and the gripper pose.
For R2R, we provide 1k trajectories in total in each domain. 
For V2V-Reach, we provided 300 trajectories for each task (i.e., each goal color).
For V2V-Open, we provided 300 trajectories for Window-Close in total, that is, 75 trajectories for each. In adaptation, we provide 300 trajectories of Window-Open.

\subsection{Architectures and Traning Details of \ours}
\label{app:plp_detail}
Our policy architecture is a simple multilayer perceptron (MLP).
The state encoder, common policy, and decoder consist of three, four, and three hidden layers, respectively, 
with 192 units each and GELU activations \citep{hendrycks2016gaussian}. 
The final layer of the decoder uses Tanh activation.
The dimension of the latent representations is also set to 192.
For the visual inputs in V2V, we used a convolutional network with CoordConv \citep{liu2018intriguing} of channels (64, 64, 128, 128, 128), kernel size $=3$, stride $=2$, and padding $=1$, followed by a single linear layer that projects the output into a single vector with 1024 dimensions.
To keep visual information that is irrelevant for proxy tasks but necessary for the target task, we add an image reconstruction loss to the objective.

We optimized our objective with AdamW optimizer \citep{loshchilov2018decoupled}.
In the Maze environments, we set the learning rate to 5e-4 and the batch size to 256, and trained the model for 20 epochs. 
In the adaptation phase, we used about 400 trajectories towards an unseen goal to update the common policy for 50 epochs.
For R2R-Lift, we set the learning rate to 1e-4 and the batch size to 512, and trained the model for 100 epochs.
For V2V, we set the learning rate to 5e-4 and the batch size to 64, and trained the model for 80 epochs.
As mentioned in Section \ref{sec:adaptation}, we mixed the dataset for the alignment so that domain-specific components does not leak into the common policy. 
We randomly selected data from the alignment dataset of twice the size of the adaptation dataset in P2P and R2R, and that of the same size for P2A and V2V.
We also found that adding weight decay stabilizes the alignment and the performance of P2P and P2A. We set the coefficient to 1e-5 in these environments. 
We did not use weight decay in R2R and V2V.
For a kernel for MMD, we use Gaussian kernel: $f(a, b) = \exp (- \|a - b\|^2 / h)$. $h$ is set to 1 in all experiments.
Since we observed that it decreased the scale of representation $s_z$ and deteriorated the performance, 
we normalized the distance $\|a-b\|$ by the average pairwise distance in a batch to mitigate the issue. Similar normalization by medians is applied in previous literature as well \citep{gretton2012kernel}.
For the TCC part, we reduced the batch size to 64, 32, and 8 in the Maze, R2R, and V2V environments respectively to reduce training time. 
The number of gradient steps for the TCC part is kept the same as the other parts.
Additionally, to alleviate the difficulty of classification in TCC, we decimated the trajectories by selecting one out of every 16 frames.
Regarding the coefficients in the objective \ref{eq:final}, we set 
$\lambda_\text{MMD} = 0.1, \lambda_\text{TCC} = 0.2$ for P2P, P2A, and R2R, and
$\lambda_\text{MMD} = 0.1, \lambda_\text{TCC} = 0.1$ for V2V environments.
In \ours-disc for the ablation, where we utilized the GAN-like discriminative loss instead of MMD regularization, we introduced a discriminator network with four hidden layers. The coefficient for the discriminative loss was set to 0.1 for P2P, P2A, and R2R, while we used 0.01 for V2V experiments.
The training time was about an hour for the maze environments and R2R,
four hours for R2R-Lift, and three hours for the V2V environments with a single GPU.

\subsection{Baselines}
\label{app:baselines}
For \gama, we re-implemented the algorithm by referring to the original paper and an official implementation (\url{https://github.com/ermongroup/dail}).
When we found discrepancies between the paper and the implementation, we followed the descriptions in the paper.
We swept the adversarial coefficient from 0.01 to 10, the learning rate from 1e-4 to 1e-3, and selected 0.5 and 1e-4, respectively.

For \cdil, we re-implemented the algorithm based on the paper.
First, we trained the temporal position models in both the source and target domains.
Subsequently, we trained all models using cycle consistency loss, adversarial loss, and temporal position loss between domains.
At the same time, we performed inference task adaptation using data from the target task in the source domain.
Next, we trained the inverse dynamics model in the target domain.
Once the target task data in the source domain was converted into the target domain, 
the inverse dynamics model was used to compute the target task actions in the target domain.
Finally, the final policy was obtained through behavioral cloning using this data.

For the demonstration-conditioned model (\cond), we used a Transformer \citep{vaswani2017attention}-based architecture to process sequences of observations and actions.
We fed the demonstrations and observation history as they were without thining-out timesteps. 
The maximum sequence length was 250 and 400 for the umaze and medium maze in P2P, 350 and 600 in P2A, and 200 in R2R and V2V, respectively.
The model had three encoder layers and three decoder layers, each with 256 units and eight heads. 
The dropout rate was set to 0.2. 
The activation function was GELU and the normalization was applied before the residual connections.
We set the batch size to 64, the learning rate to 1e-3, and trained the model for 500 epochs in each environment in Maze environments and R2R. In V2V, we set the batch size to 16.

\end{document}